\documentclass{article}

 \usepackage[main, final]{neurips_2026}

\usepackage[utf8]{inputenc} 
\usepackage[T1]{fontenc}    
\usepackage{hyperref}       
\hypersetup{hidelinks}
\usepackage{url}            
\usepackage{booktabs}       
\usepackage{amsfonts}       
\usepackage{nicefrac}       
\usepackage{microtype}      
\usepackage{xcolor}         

\definecolor{Coral}{rgb}{1, 0.47, 0.24}

\usepackage{booktabs}
\usepackage{multirow}
\usepackage{graphicx}
\usepackage{array}
\usepackage{gensymb}  
\usepackage{fancyvrb}
\usepackage{lipsum}
\usepackage{xspace}
\usepackage{pifont}

\usepackage{booktabs} 
\usepackage[T1]{fontenc}    

\usepackage{url}            
\usepackage{booktabs}       
\usepackage{amsfonts}       
\usepackage{nicefrac}       
\usepackage{microtype}      
\usepackage{xcolor}         
\usepackage{graphicx} 
\usepackage{threeparttable}
\usepackage{multirow}
\usepackage{mathtools}
\usepackage{tabularx}
\usepackage{color}
\usepackage{colortbl}
\usepackage{amsthm}
\usepackage{caption}
\usepackage{marvosym}
\usepackage{enumitem}
\usepackage{bm}
\usepackage{wrapfig}

\usepackage{gensymb}
\usepackage{makecell}
\usepackage{svg} 
\usepackage{float}
\usepackage{subcaption}
 
\usepackage{multicol}
\usepackage{xspace}

\usepackage{pgfplots}
\pgfplotsset{compat=1.18}
\usetikzlibrary{backgrounds}

\usepackage[ruled]{algorithm2e} 

\title{PLANING: A Loosely Coupled Triangle-Gaussian Framework for Streaming 3D Reconstruction}

\author{
Changjian Jiang\textsuperscript{1,2,3*}\quad
Kerui Ren\textsuperscript{4,2*} \quad
Xudong Li\textsuperscript{5,2} \quad
Kaiwen Song\textsuperscript{6,2} \quad
Guanghao Li\textsuperscript{7} \\ 
\textbf{Linning Xu\textsuperscript{8,2}} \quad 
\textbf{Tao Lu\textsuperscript{2}} \quad 
\textbf{Junting Dong\textsuperscript{2}} \quad
\textbf{Yu Zhang\textsuperscript{1$\dagger$}} \quad 
\textbf{Yu Feng\textsuperscript{9}} \quad
\textbf{Bo Dai\textsuperscript{3,10}} \quad
\textbf{Mulin Yu\textsuperscript{2$\dagger$}} \quad
\vspace{4mm} \\
\textsuperscript{1}Zhejiang University\quad 
\textsuperscript{2}Shanghai Artificial Intelligence Laboratory\quad 
\textsuperscript{3}The University of Hong Kong
\\
\textsuperscript{4}Shanghai Jiao Tong University\quad 
\textsuperscript{5}Northwestern Polytechnical University\quad \\
\textsuperscript{6}University of Science and Technology of China\quad 
\textsuperscript{7}Fudan University\quad \\
\textsuperscript{8}The Chinese University of Hong Kong\quad 
\textsuperscript{9}Zhejiang University of Technology \\ 
\textsuperscript{10}HKU Shanghai Intelligent Computing Research Center \\
\small\texttt{changjianjiang01@gmail.com, zhangyu80@zju.edu.cn, yumulin@pjlab.org.cn}
}

\newcommand{\modelname}{PLANING}

\begin{document}

\maketitle

{
\begin{center}
    \centering
    \captionsetup{type=figure}
    \includegraphics[width=\textwidth]{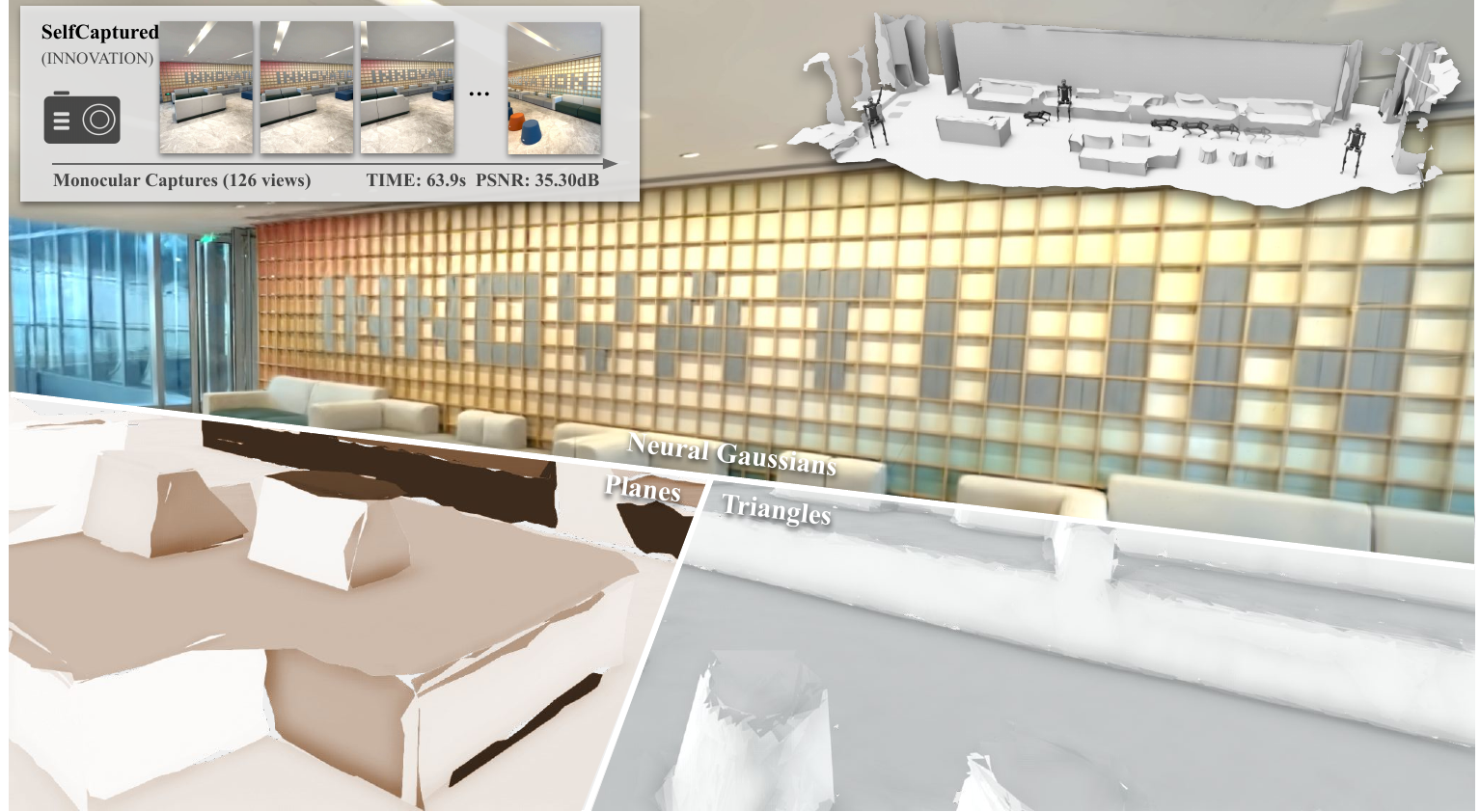}
    \captionof{figure}
    {
    \modelname~introduces a loosely coupled triangle-Gaussian representation for streaming 3D reconstruction, balancing geometric accuracy, high-fidelity rendering, and computational efficiency. Compared with 2DGS-based representations, explicit edge-preserving triangle primitives provide sharper structural boundaries and more stable geometric learning, leading to improved geometric reconstruction quality. The extracted compact planar structures further support high-performance simulation environments for locomotion training in embodied AI. Project page: 
\href{https://city-super.github.io/PLANING/}{\textcolor{magenta}{\textbf{https://city-super.github.io/PLANING/}}}.
    }
    \label{fig:teaser}
\end{center}
}

{\let\thefootnote\relax\footnotetext{{$^*$ Equal contribution.} $^\dagger$ Corresponding author.}}

\begin{abstract}
Streaming reconstruction from monocular image sequences remains challenging, as existing methods typically favor either high-quality rendering or accurate geometry, but rarely both.
We present \modelname, an efficient on-the-fly reconstruction framework built on a hybrid representation that loosely couples explicit geometric primitives with neural Gaussians, enabling geometry and appearance to be modeled in a decoupled manner.
This decoupling supports an online initialization and optimization strategy that separates geometry and appearance updates, yielding stable streaming reconstruction with substantially reduced structural redundancy.
Despite operating online, \modelname~improves dense mesh Chamfer-L2 by 18.52\% over the offline baseline PGSR, surpasses the streaming baseline ARTDECO by 1.31 dB in PSNR on average across the six evaluated datasets, and reconstructs ScanNetV2 scenes in 2.1 minutes, over 5$\times$ faster than offline 2D Gaussian Splatting while matching per-scene optimization quality.
Beyond reconstruction quality, the structural clarity and computational efficiency of \modelname~make it well suited for a broad range of downstream applications, such as enabling large-scale scene modeling and simulation-ready environments for embodied AI.
\end{abstract}

\section{Introduction}
\label{sec:intro}
3D scene reconstruction is a core capability for embodied intelligence, autonomous driving, and AR/VR, providing the spatial understanding required for perception and interaction~\cite{yang2024physcene,qi2024air}. While offline reconstruction methods following a capture-then-process paradigm have reached a high level of maturity, their reliance on time-intensive post-processing limits scalability and responsiveness in time-critical scenarios. This has driven a growing demand for on-the-fly reconstruction frameworks.

A central challenge in on-the-fly 3D reconstruction is a scene representation that jointly achieves high geometric accuracy and computational efficiency. Recently, 3D Gaussian Splatting (3DGS)~\cite{kerbl20233d} has emerged as a compelling explicit representation, offering high visual fidelity with efficient rendering, and has therefore been widely adopted in streaming reconstruction methods~\cite{meuleman2025fly,li2025artdeco,cheng2025outdoor,keetha2024splatam,matsuki2024gaussian,ren2026m}.
Despite their success, existing streaming 3DGS-based methods share a fundamental limitation: \textit{the absence of explicit, compact, and stable geometry.} While Gaussian primitives excel at appearance modeling, their ambiguous structural boundaries hinder stable surface reconstruction and geometric consistency. To compensate, these methods rely on a large number of primitives, leading to significant redundancy, increased computational cost, and limited scalability in streaming settings.

\begin{wrapfigure}{r}{0.35\linewidth}
\vspace{-16pt}
\centering
\includegraphics[width=1\linewidth]{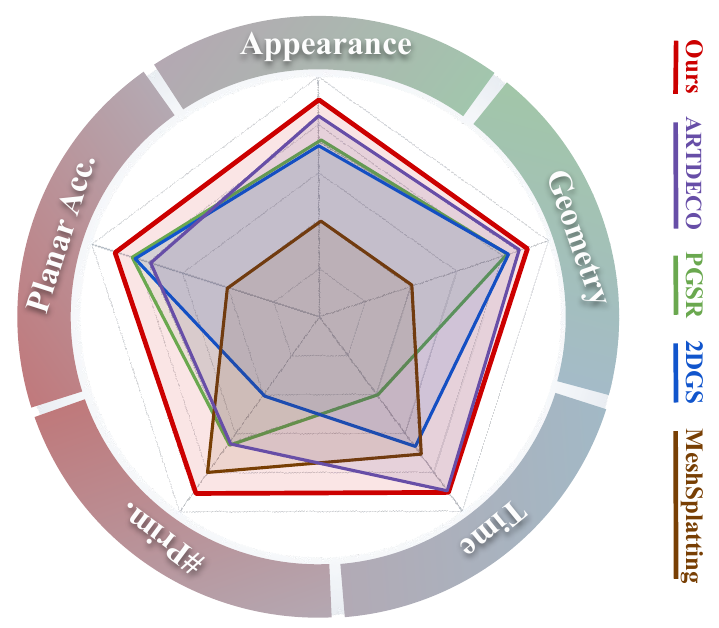}
\vspace{-10pt}
\caption{
\modelname~consistently outperforms existing streaming and per-scene reconstruction methods across geometry accuracy, rendering quality, computational efficiency, and primitive count, while maintaining clear and well-structured planar geometry.
}
\label{fig:pie}
\vspace{-14pt}
\end{wrapfigure}

To address these challenges, (1) we develop a hybrid representation that decouples geometry from appearance for efficient reconstruction and high-fidelity rendering. For \emph{geometry}, we introduce learnable triangle primitives with explicit edge definitions, enabling sharper structural boundaries and more stable geometric learning than 2DGS-based representations, which leads to improved geometric reconstruction quality. For \emph{appearance modeling}, we adopt a neural-Gaussian formulation inspired by Scaffold-GS~\cite{lu2024scaffold}, where Gaussian attributes are decoded from fused triangle and Gaussian features to encourage local rendering consistency. This design establishes a complementary interaction: triangles serve as stable structural anchors, while rendering gradients propagated through neural Gaussians refine the underlying geometry in a controlled manner.
Building upon this representation, (2) we introduce \modelname, an efficient streaming framework for monocular 3D reconstruction. Our framework leverages feed-forward geometric priors for robust pose estimation and stable scene initialization, while photometric and spatial filtering suppress redundant primitives. Furthermore, global map update enhances global geometric consistency during streaming reconstruction.

Extensive experiments across diverse indoor and outdoor benchmarks demonstrate that our method outperforms state-of-the-art approaches in geometric accuracy, rendering quality, training efficiency, and primitive count, as illustrated in Fig.~\ref{fig:teaser} and Fig.~\ref{fig:pie}. 
By preserving salient structures and suppressing redundant geometry, our representation enables the extraction of compact and consistent 3D planes with significantly reduced triangle counts. This compact geometric representation is particularly beneficial for large-scale reconstruction and globally consistent pose estimation. Furthermore, the structural clarity and efficiency of our model make it well suited for simulation-ready scene modeling, such as supporting local motion policy training in embodied AI.
Our main contributions can be summarized as follows:

\noindent\textbf{(1)} We develop \textbf{a hybrid representation} that decouples geometry and appearance by combining explicit triangle primitives with neural Gaussians, enabling compact and stable geometry learning and high-fidelity rendering.

\noindent\textbf{(2)} We develop \textbf{an efficient on-the-fly monocular reconstruction framework} that leverages the proposed representation together with streaming-aware initialization and global map update.

\noindent\textbf{(3)} We demonstrate \textbf{state-of-the-art performance} in both geometric accuracy and rendering quality across diverse indoor and outdoor benchmarks, and showcase the versatility of our approach for downstream tasks including plane-guided pose refinement, large-scale scene reconstruction, and simulation-ready environments for embodied AI.
\section{Related Work} 
\label{sec:related_works}

\textbf{3D Reconstruction and SLAM.}
Reconstructing 3D geometry from multi-view observations is a fundamental problem spanning offline reconstruction and online SLAM. Classical pipelines~\cite{schonberger2016pixelwise,mur2017orb,campos2021orb,qin2018vins} rely on calibrated images to recover camera poses and sparse geometry, but often lack the fidelity required for high-quality rendering. Neural rendering methods, notably NeRF~\cite{mildenhall2021nerf}, model scenes as continuous fields and enable photorealistic synthesis, yet their implicit formulation and costly volumetric sampling limit scalability and real-time applicability. 

To improve efficiency, 3D Gaussian Splatting (3DGS)~\cite{kerbl20233d} introduces explicit primitives with rasterization-based rendering, enabling real-time reconstruction and making it attractive for streaming scenarios~\cite{matsuki2024gaussian,keetha2024splatam,ha2024rgbd}. Recent works integrate 3DGS into SLAM pipelines by jointly optimizing camera poses and scene representations via rendering supervision, or by adopting incremental updates with lightweight bundle adjustment~\cite{zhang2025hi,meuleman2025fly}. Despite these advances, existing approaches struggle to simultaneously achieve robustness, geometric accuracy, and efficiency, particularly in monocular and long-sequence settings.

An alternative paradigm leverages feed-forward models pretrained on large-scale datasets~\cite{murai2025mast3r,wang2025pi3,wang2025vggt,lin2025depth,zhuo2025streaming} to directly infer scene structure with or without known poses. While these methods offer strong generalization and fast inference, they typically lag behind optimization-based approaches in reconstruction accuracy and global consistency. Overall, achieving a unified representation that balances geometric fidelity, efficiency, and robustness remains an open challenge.

\textbf{3DGS Variants.}
Various extensions have explored alternative primitives to better align with scene geometry. 2DGS~\cite{huang20242d}, GSS~\cite{dai2024high}, and Quadratic Gaussian Splatting~\cite{zhang2025quadratic} replace anisotropic Gaussians with ellipsoidal or quadric forms for superior surface alignment. Other works incorporate explicit geometric elements, such as the 3D convexes~\cite{held20253d} and triangles~\cite{jiang2026halogs,held2025meshsplatting,burgdorfer2025radiant}, to compactly model hard-edged scenes. Similarly, PlanarSplatting~\cite{tan2025planarsplatting} utilizes rectangular primitives to achieve structured and efficient indoor planar reconstructions.
Despite these advances, single-representation methods often struggle to balance geometric precision with rendering fidelity. To bridge this gap, recent dual-branch approaches such as GSDF~\cite{yu2024gsdf} and 3DGSR~\cite{lyu20243dgsr} integrate neural signed distance fields (SDFs) with 3DGS. While this enables partial geometry–appearance decoupling, it introduces significant computational overhead and optimization complexity. Alternatively, 3D-GES~\cite{ye2025gaussian} adopts a bi-scale formulation using 2D surfels for coarse structure and 3D Gaussians for fine detail. However, this design primarily targets appearance enhancement rather than achieving a principled, explicit decoupling of geometry and appearance.
\section{Method}
\label{sec:method}

In this section, we first introduce our dual scene representation that combines learnable triangles with neural Gaussians (Sec.~\ref{Decoupled Scene Representations}). We then describe how we adapt this representation into an on-the-fly reconstruction framework, achieving both efficiency and high-quality 3D reconstruction (Sec.~\ref{sec:streaming_reconstruction_framework}).

\subsection{Loosely-coupled Triangle-Gaussian Representation}
\label{Decoupled Scene Representations}
We first detail the triangle primitives and our differentiable rasterizer. Subsequently, we explain the interaction between neural Gaussians and their corresponding triangles.

\subsubsection{Learnable Triangles for Geometry}
We propose triangle primitives based on a vertex-based formulation and a differentiable rasterizer.

\textbf{Vertex-Based Primitive Definition.}
We parameterize each triangle primitive by its three learnable vertices $\{\mathbf{p}_0, \mathbf{p}_1, \mathbf{p}_2\}$.
To facilitate efficient and differentiable rendering, we define a local coordinate frame for each triangle: 
\begin{equation}
\begin{aligned}
& \mathbf{t}_u = \dfrac{\mathbf{p}_0 - \bm{\mu}}{\left\|\mathbf{p}_0 - \bm{\mu}\right\|_2},   
\mathbf{t}_v = \mathbf{n} \times \mathbf{t}_u ,  
s_u = \left\|\mathbf{p}_0 - \bm{\mu}\right\|_2,   \\
& s_v = \left| \mathbf{t}_v \cdot (\mathbf{p}_1 - \bm{\mu})\right|, 
\mathbf{n}= \dfrac{(\mathbf{p}_1 - \mathbf{p}_0) \times (\mathbf{p}_2 - \mathbf{p}_0)}
{\left\|(\mathbf{p}_1 - \mathbf{p}_0) \times (\mathbf{p}_2 - \mathbf{p}_0)\right\|_2},
\end{aligned}
\label{eq:local_triangle_frame}
\end{equation}
where the barycenter $\bm{\mu}$ is set as the origin of the local frame.
Under this construction, the three vertices can be expressed in the local tangent plane as $\{\mathbf{p}_0^{\prime}, \mathbf{p}_1^{\prime}, \mathbf{p}_2^{\prime}\} = \{(0, 1)^T, (a, 1)^T, (-1-a, -1)^T\}$, where $a = \mathbf{t}_u \cdot (\mathbf{p}_1 - \bm{\mu})$ is the only degree of freedom in the local frame.

Following 3D Convex Splatting (3DCS)~\cite{held20253d}, we further introduce two learnable triangle-wise parameters, $\delta > 0$ and $\sigma > 0$, to control edge sharpness and boundary smoothness.
Each triangle is also associated with a learnable opacity parameter $\alpha$, analogous to 3DGS~\cite{kerbl20233d}.

\begin{figure*}[t!]
\centering
\includegraphics[width=\linewidth]{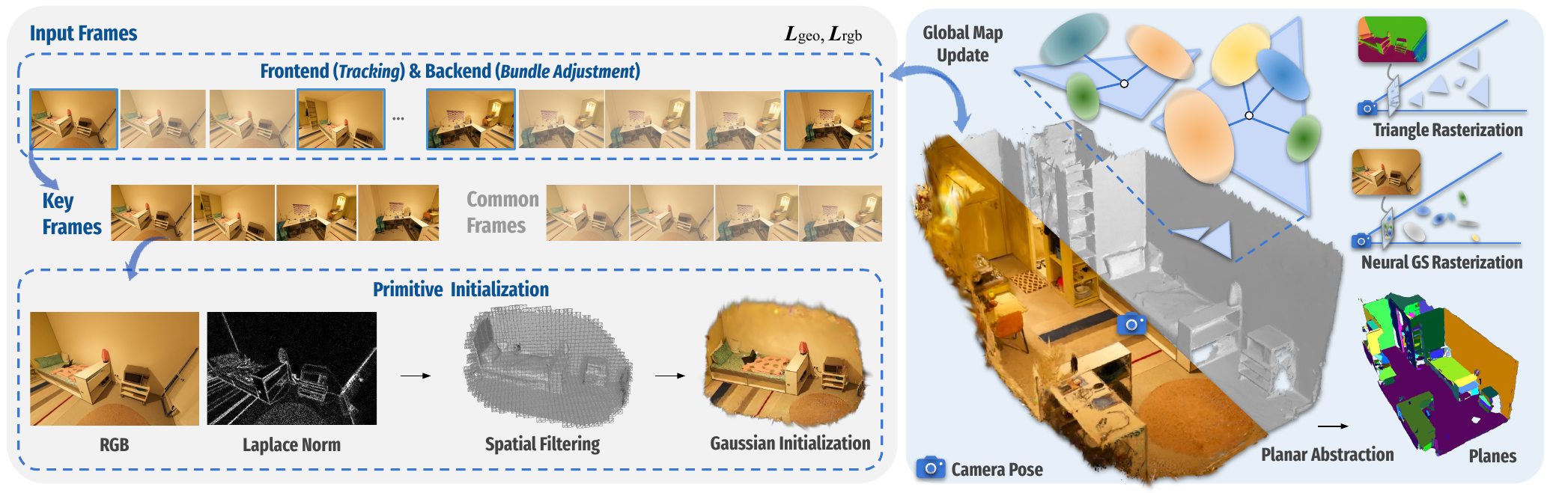}
\caption{
\textbf{Pipeline of \modelname.} \modelname~adopts a hybrid representation in which triangles explicitly model scene geometry, while neural Gaussians decoded from these triangles render appearance. Built upon this representation, we develop a streaming reconstruction framework for unposed monocular image sequences, consisting of a frontend for camera tracking, a backend for global pose optimization, and a mapper for scene reconstruction. The mapper further incorporates an efficient primitive initialization strategy to suppress redundancy. The reconstructed triangle soup additionally enables compact planar abstraction for downstream applications.
}
\label{fig:pipeline}
\end{figure*}

\textbf{Differentiable Triangle Rasterizer.} 
We implement an efficient differentiable triangle rasterizer that enables direct supervision of triangles using prior normals and depths.
To obtain unbiased depth rendering, we adopt an explicit ray-triangle intersection strategy, similar in spirit to 2DGS~\cite{huang20242d}. We further introduce the edge-preserving contribution function as:
\begin{equation}
w(\hat{\mathbf{x}})=
\mathrm{Sigmoid}\left(-\sigma\log\left(\sum_{j=0}^2\exp\left(\delta~
\mathrm{dist}(\hat{\mathbf{x}}, \mathbf{e}_j)\right)\right)\right)\alpha, 
\label{eq:contribution}
\end{equation}
where $\mathrm{dist}(\hat{\mathbf{x}}, \mathbf{e}_j)$ denotes the distance from the intersection point $\hat{\mathbf{x}}$ to the $j$-th triangle edge in the local tangent plane.
Thanks to the local frame parameterization, these distances can be computed analytically as:
\begin{equation}
\begin{aligned}
\left\{
\begin{array}{l}
\mathrm{dist}(\hat{\mathbf{x}}, \mathbf{e}_0) = u+(1-a)v-1, \\[3pt]
\mathrm{dist}(\hat{\mathbf{x}}, \mathbf{e}_1) = -2u+(2a+1)v-1, \\[3pt]
\mathrm{dist}(\hat{\mathbf{x}}, \mathbf{e}_2) = u+(-2-a)v-1,
\end{array}
\right.
\end{aligned} 
\label{eq:local_triangle_dis}
\end{equation}
where $\hat{\mathbf{x}} = (u, v)^T$. This closed-form formulation significantly simplifies both forward evaluation and gradient propagation.
Notably,  our contribution computation differs from 3DCS, where contributions are computed directly on the image plane rather than in the local surface domain.

Finally, triangles are rendered into depth and normal maps using front-to-back alpha compositing:
\begin{equation}
\mathbf{N}(\mathbf{x}) = \sum_{i=1}^N \mathbf{n}_i w\left(\hat{\mathbf{x}}_i\right)  \prod_{j=1}^{i-1} \left(1 - w\left(\hat{\mathbf{x}}_j\right) \right), \quad 
\mathbf{D}(\mathbf{x}) = \sum_{i=1}^N d_i w\left(\hat{\mathbf{x}}_i\right)  \prod_{j=1}^{i-1} \left(1 - w\left(\hat{\mathbf{x}}_j\right) \right),
\label{eq:NandD}
\end{equation}
where $d_i$ denotes the distance from the $i$-th intersection point to the pixel. The $N$ ordered intersection points $\{\hat{\mathbf{x}}_i\}$ between the triangles and pixel $\mathbf{x}$ are computed using our custom CUDA-based rasterizer. To enable accurate differentiable rendering of triangles, we define a new criterion for visibility determination and design a triangle-subdivision-based primitive depth sorting algorithm in the rasterizer to address rendering issues introduced by the edge-preserving contribution function
. The detailed forward rendering pipeline is described in the Appendix~\ref{sec:appendix_rasterizer}.

\subsubsection{Neural Gaussians for Appearance Modeling}
To achieve a decoupled yet consistent representation of geometry and appearance, we introduce neural Gaussians to flexibly encode view-dependent appearance. Inspired by Scaffold-GS~\cite{lu2024scaffold}, neural Gaussians are anchored to the triangles and used for appearance. Specifically, each learnable triangle is associated with a context feature $\mathbf{f}_{\text{t}} \in \mathbb{R}^{24}$. Each Gaussian is parameterized by a learnable position offset $\mathbf{o}_{\text{g}}\in\mathbb{R}^3$, spherical harmonics (SH) coefficients, an opacity $\alpha_{\text{g}}\in\mathbb{R}$, a base scale $\mathbf{s}_{\text{g}}\in\mathbb{R}^3$, a base quaternion $\mathbf{q}_{\text{g}}\in\mathbb{R}^4$, and an individual feature $\mathbf{f}_{\text{g}} \in \mathbb{R}^{8}$. 

During rendering, the position of each Gaussian $\bm{\mu}_{\text{g}} = \mathbf{o}_{\text{g}} + \bm{\mu}_{\text{t}}$,
where $\bm{\mu}_{\text{t}}$ denotes the barycenter of the associated triangle. Then we predict the final scale $\mathbf{s}=\mathbf{s}_{\text{g}} \odot \mathrm{MLP}_s(\mathbf{f}_{\text{t}} \oplus \mathbf{f}_{\text{g}})$ and rotation $\mathbf{q} = \mathrm{\phi}(\mathbf{q}_{\text{g}}\odot\mathrm{MLP}_q(\mathbf{f}_{\text{t}} \oplus \mathbf{f}_{\text{g}}))$,
where $\odot$ denotes element-wise multiplication, $\oplus$ denotes feature concatenation and $\mathrm{\phi}(\cdot)$ denotes $\ell_2$ normalization to ensure valid rotation quaternions. Through this design, geometry and appearance are represented in a consistent manner. 
Notably, each triangle hosts a flexible number of Gaussians, enabling the representation to adapt to local scene details.

\subsection{Streaming Reconstruction Framework}
\label{sec:streaming_reconstruction_framework}
\subsubsection{Overview}
As shown in Fig.~\ref{fig:pipeline}, we build a streaming monocular reconstruction framework upon the proposed hybrid representation. Following~\cite{li2025artdeco}, the system consists of a frontend for camera tracking, a backend for global pose optimization, and a mapper for scene reconstruction.

The frontend processes incoming frames in a streaming manner to estimate camera poses, select keyframes, and predict dense point maps using feed-forward models~\cite{leroy2024grounding}. The backend performs loop closure detection~\cite{wang2025pi3} and global bundle adjustment~\cite{murai2025mast3r} to improve global pose consistency. Given the optimized poses and geometric priors from the backend, the mapper incrementally reconstructs scene geometry and appearance using the proposed triangle-Gaussian representation. The resulting map is continuously available throughout the streaming process.

To improve efficiency and geometric consistency, we introduce redundancy-aware primitive initialization and global map update during streaming reconstruction. After all input frames have been processed, we apply a final global refinement to the existing map to improve global consistency and rendering quality. 
Following the final refinement, planar structures can be directly extracted from the triangle soup via a coarse-to-fine plane extraction algorithm. Furthermore, our framework supports dense mesh reconstruction through depth fusion. 
Details are provided in Appendices~\ref{sec:streaming_global_refinement}--\ref{sec:implementation_details}.

\subsubsection{Primitive Initialization}

Upon receiving a keyframe from the backend, the framework initializes new primitives in regions with insufficient geometric coverage or high reconstruction error. To identify informative regions, we compute an insertion probability map based on the discrepancy between the rendered image and the observation using a Laplacian-of-Gaussian (LoG) operator~\cite{meuleman2025fly}. New primitives are instantiated only when the insertion probability exceeds a stochastic threshold with upper bound $\tau_a$, encouraging the model to focus on high-frequency and under-reconstructed regions. Implementation details are provided in Appendix~\ref{sec:prim_init_supp}.

\textbf{Spatial Filtering.} To suppress structural redundancy and maintain map compactness, we further apply a depth-adaptive spatial filtering strategy. For each candidate primitive, we compute its back-projected 3D center and discard it if nearby primitives already exist within a local vicinity. The vicinity radius is adaptively scaled according to the observation depth:
\begin{equation}
    V(d_i) = V_{\min} + (V_{\max} - V_{\min}) 
    \left( \frac{d_i - d_{\min}}{d_{\max} - d_{\min}} \right)^p,
\label{eq:vicinity_radius}
\end{equation}
where $d_i$ denotes the observation depth and 
$\{V_{\min}, V_{\max}, d_{\min}, d_{\max}, p\}$ are hyperparameters controlling the spatial extent. This strategy effectively prevents excessive primitive growth in already-reconstructed regions. Once a candidate location is selected, a triangle primitive is initialized using the local depth and normal priors. 
The triangle scale is determined according to the local geometric structure, while its orientation is aligned with the predicted surface normal.

\textbf{Gaussian Initialization.} To model appearance details, neural Gaussians are subsequently initialized around triangle barycenters. The number of Gaussians is adaptively determined based on local image complexity, enabling higher representational capacity in texture-rich regions. Gaussian attributes
are initialized using local geometric and photometric observations.

\begin{table*}[!t]
\centering
\renewcommand{\arraystretch}{1.15}
\setlength{\tabcolsep}{1pt}
\caption{\textbf{Quantitative comparison of planar reconstruction.} We evaluate the geometric and planar metrics on the ScanNet++, ScanNetV2, and FAST-LIVO2 datasets. Ours achieves top-tier performance in most categories while significantly reducing primitive count and runtime (reported in minutes).}
\label{tab:plane_recon_comp}
\resizebox{\linewidth}{!}{
\begin{tabular}{l|cc|ccc|c|c|cc|ccc|c|c|cccc|c|c}
\toprule
\multirow{3}{*}{Method} & \multicolumn{7}{c|}{ScanNet++} & \multicolumn{7}{c|}{ScanNetV2} & \multicolumn{6}{c}{FAST-LIVO2} \\
& \multicolumn{2}{c|}{Geometry} & \multicolumn{3}{c|}{Planar} & \multirow{2}{*}{\makecell{Time}} & \multirow{2}{*}{\#Prim.} & \multicolumn{2}{c|}{Geometry} & \multicolumn{3}{c|}{Planar} & \multirow{2}{*}{\makecell{Time}} & \multirow{2}{*}{\#Prim.} &\multicolumn{4}{c|}{Geometry}& \multirow{2}{*}{\makecell{Time}} & \multirow{2}{*}{\#Prim.}\\
& Ch-L2$\downarrow$ & F-score$\uparrow$ & Fidelity$\downarrow$ & Acc$\downarrow$ & Ch-L2$\downarrow$ & & & Ch-L2$\downarrow$ & F-score$\uparrow$ & Fidelity$\downarrow$ & Acc$\downarrow$ & Ch-L2$\downarrow$ & & & Acc$\downarrow$ & Comp$\downarrow$ & Ch-L2$\downarrow$ & F-score$\uparrow$ & & \\
\midrule 
2DGS$^\dagger$ & 3.89 & 81.64 & \cellcolor{yellow!40}8.16 & \cellcolor{orange!40}7.19 & \cellcolor{yellow!40}7.67 & 16.1 & 415.3k & 6.48 & 53.73 & 15.56 & \cellcolor{orange!40}8.12 & 11.84 & 10.9 & 1196.8k & \cellcolor{yellow!40}14.11 & \cellcolor{orange!40}48.17 & \cellcolor{orange!40}53.45 & \cellcolor{yellow!40}60.47 & 35.8 & 3197.0k \\
PGSR$^\dagger$ & \cellcolor{yellow!40}3.87 & \cellcolor{yellow!40}81.98 & \cellcolor{orange!40}7.44 & \cellcolor{yellow!40}7.23 & \cellcolor{orange!40}7.33 & 31.2 & \cellcolor{yellow!40}353.4k & 6.59 & 54.28 & 15.88 & \cellcolor{yellow!40}8.46 & 12.17 & 21.3 & 629.1k & \cellcolor{orange!40}13.95 & \cellcolor{yellow!40}49.16 & \cellcolor{yellow!40}54.13 & \cellcolor{orange!40}60.75 & \cellcolor{yellow!40}25.5 & \cellcolor{yellow!40}1065.8k \\
MeshSplatting$^\dagger$ & 9.13 & 47.19 & 37.87 & 10.71 & 24.29 & 38.5 & 1825k & 11.15 & 30.73 & 40.16 & 11.45 & 25.81 & 9.7 & \cellcolor{yellow!40}291.3k & 14.52 & 66.68 & 62.97 & 47.31 & 26.3 & 2505.1k \\
\midrule
AirPlanes & 25.19 & 19.21 & 47.10 & 25.97 & 36.53 & \cellcolor{red!40}3.7 & / & \cellcolor{yellow!40}6.34 & \cellcolor{yellow!40}55.33 & \cellcolor{red!40}9.68 & 8.90 & \cellcolor{orange!40}9.29 & 3.5 & / & - & - & - & - & - & - \\
PlanarSplatting & 7.27 & 49.78 & 9.64 & 13.35 & 11.50 & 8.8 & \cellcolor{red!40}1.0k & 6.54 & 51.67 & \cellcolor{orange!40}9.72 & 10.77 & \cellcolor{yellow!40}10.24 & \cellcolor{yellow!40}3.1 & \cellcolor{red!40}1.76k & - & -& - & - & - & - \\
\midrule
ARTDECO & \cellcolor{orange!40}3.82 & \cellcolor{orange!40}83.08 & 15.84 & 7.92 & 11.88 & \cellcolor{yellow!40}5.6 & 478.3k & \cellcolor{orange!40}6.05 & \cellcolor{orange!40}57.58 & 19.62 & 8.73 & 14.18 & \cellcolor{orange!40}2.2 & 621.5k & 14.17 & 63.63 & 57.19 & 54.23 & \cellcolor{orange!40}6.5 & \cellcolor{orange!40}501.6k \\ 
\midrule
Ours & \cellcolor{red!40}3.53 & \cellcolor{red!40}86.88 & \cellcolor{red!40}7.24 &\cellcolor{red!40} 6.95 & \cellcolor{red!40}7.09 & \cellcolor{orange!40}5.5 & \cellcolor{orange!40}61.6k & \cellcolor{red!40}5.68 & \cellcolor{red!40}62.15 & \cellcolor{yellow!40}10.55 & \cellcolor{red!40}7.58 & \cellcolor{red!40}9.07 & \cellcolor{red!40}2.1 & \cellcolor{orange!40}56.1k & \cellcolor{red!40}12.58 & \cellcolor{red!40}30.60 & \cellcolor{red!40}36.77 & \cellcolor{red!40}65.89 & \cellcolor{red!40}3.6 & \cellcolor{red!40}101.4k  \\
\bottomrule
\end{tabular}}
\begin{tablenotes}
\footnotesize
\item ~/: w/o explicit geometric primitives, ~–: beyond the scope (indoor scenes) of the method, ~$^\dagger$: offline per-scene baselines are augmented with the same geometric-prior supervision as \modelname.
\end{tablenotes}
\end{table*}

\begin{table*}[t!]
\centering
\renewcommand{\arraystretch}{1.15}
\setlength{\tabcolsep}{1pt}
\caption{\textbf{Quantitative comparison of appearance rendering.} We evaluate the rendering quality metrics across six diverse indoor and outdoor datasets. Our method achieves state-of-the-art performance in most categories while significantly reducing the runtime (reported in minutes).}
\label{tab:rendering_comp}
\resizebox{\linewidth}{!}{
\begin{tabular}{l|ccc|ccc|ccc|ccc|ccc|ccc|c}
\toprule
\multirow{2}{*}{Method} & \multicolumn{3}{c|}{ScanNetV2} & \multicolumn{3}{c|}{VR-NeRF} & \multicolumn{3}{c|}{ScanNet++} & \multicolumn{3}{c|}{Waymo} & \multicolumn{3}{c|}{FAST-LIVO2} & \multicolumn{3}{c|}{KITTI} & \multirow{2}{*}{\makecell{Time}} \\
& PSNR$\uparrow$ & SSIM$\uparrow$ & LPIPS$\downarrow$ 
& PSNR$\uparrow$ & SSIM$\uparrow$ & LPIPS$\downarrow$ 
& PSNR$\uparrow$ & SSIM$\uparrow$ & LPIPS$\downarrow$ 
& PSNR$\uparrow$ & SSIM$\uparrow$ & LPIPS$\downarrow$ 
& PSNR$\uparrow$ & SSIM$\uparrow$ & LPIPS$\downarrow$ 
& PSNR$\uparrow$ & SSIM$\uparrow$ & LPIPS$\downarrow$ 
& \\
\midrule 
2DGS$^\dagger$ & \cellcolor{yellow!40}27.74 & 0.873 & 0.234 & \cellcolor{yellow!40}29.61 &  \cellcolor{yellow!40}0.905 & \cellcolor{orange!40}0.200 & \cellcolor{red!40}32.00 & \cellcolor{yellow!40}0.937 & \cellcolor{red!40}0.129 & 26.97 & 0.854 & 0.324 & \cellcolor{yellow!40}29.24 & 0.867 & 0.287 & 22.29 & 0.751 & 0.338 & 31.9 \\
PGSR$^\dagger$  & 27.73 & \cellcolor{orange!40}0.880 & \cellcolor{yellow!40}0.233 & 29.37 & 0.903 & \cellcolor{yellow!40}0.201 & 31.38 & \cellcolor{yellow!40}0.937 & \cellcolor{orange!40}0.133 & \cellcolor{orange!40}27.42 & \cellcolor{yellow!40}0.865 & 0.306 & 29.22 & \cellcolor{yellow!40}0.870 & \cellcolor{yellow!40}0.280 & \cellcolor{yellow!40}22.88 &  \cellcolor{orange!40}0.785 & \cellcolor{yellow!40}0.284 & 39.9 \\
MeshSplatting$^\dagger$ & 25.64 & 0.830 & 0.351 & 25.23 & 0.819 & 0.352 & 27.71 & 0.876 & 0.294 & 23.10 & 0.781 & 0.424 & 25.78 & 0.789 & 0.397 & 17.35 & 0.551 & 0.499 & 24.6 \\
\midrule
MonoGS & 22.17 & 0.806 & 0.542 & 15.30 & 0.583 & 0.655 & 17.08 & 0.708 & 0.632 & 19.06 & 0.744 & 0.639 & 19.80 & 0.694 & 0.649 & 14.56 & 0.489 & 0.767 & 8.3 \\
S3PO-GS & 24.37 & 0.829 & 0.476 & 24.00 & 0.810 & 0.371 & 23.34 & 0.820 & 0.444 & 25.33 & 0.821 & 0.395 & 24.99 & 0.776 & 0.419 & 19.23 & 0.622 & 0.430 & 24.5  \\
\midrule
OnTheFly-NVS & 23.33 & 0.823 & 0.376 & 29.10 & 0.895 & 0.237 & 21.54 & 0.794 & 0.357 & \cellcolor{yellow!40}27.22 & 0.848 & \cellcolor{orange!40}0.300 & 21.92 & 0.735 & 0.443 & 17.17 & 0.584 & 0.427 & \cellcolor{red!40}1.3  \\
ARTDECO & \cellcolor{orange!40}28.44 & \cellcolor{yellow!40}0.877 & \cellcolor{orange!40}0.232 & \cellcolor{orange!40}30.02 & \cellcolor{orange!40}0.911 & 0.230 & \cellcolor{yellow!40}31.64 & \cellcolor{red!40}0.941 & 0.140 & 26.59 & \cellcolor{orange!40}0.869 & \cellcolor{yellow!40}0.305 & \cellcolor{orange!40}32.86 & \cellcolor{orange!40}0.926 & \cellcolor{orange!40}0.210 & \cellcolor{orange!40}22.99 & \cellcolor{yellow!40}0.777 & \cellcolor{orange!40}0.282 & \cellcolor{orange!40}6.9 \\
\midrule
Ours & \cellcolor{red!40}28.83 & \cellcolor{red!40}0.882 & \cellcolor{red!40}0.222 & \cellcolor{red!40}32.59 & \cellcolor{red!40}0.933 & \cellcolor{red!40}0.168 & \cellcolor{orange!40}31.91 & \cellcolor{red!40}0.941 & \cellcolor{orange!40}0.133 & \cellcolor{red!40}29.24 & \cellcolor{red!40}0.887 & \cellcolor{red!40}0.278 & \cellcolor{red!40}33.97 & \cellcolor{red!40}0.938 & \cellcolor{red!40}0.180 & \cellcolor{red!40}23.82 & \cellcolor{red!40}0.793 & \cellcolor{red!40}0.253 & \cellcolor{yellow!40}7.4\\
\bottomrule
\end{tabular}}
\begin{tablenotes}
\footnotesize
\item ~$^\dagger$: offline per-scene baselines are augmented with the same geometric-prior supervision as \modelname.
\end{tablenotes}
\end{table*}

\subsubsection{Training}
We supervise the triangles and Gaussians with separate geometric and appearance losses for decoupled optimization:
$\mathcal{L} = \mathcal{L}_{\text{geo}} + \mathcal{L}_{\text{rgb}}.$
For geometry, we leverage multi-view depth $\mathbf{D}_{\text{p}}$ and normal $\mathbf{N}_{\text{p}}$ priors from MASt3R~\cite{leroy2024grounding} to supervise triangles, penalizing deviations from the rendered depth $\mathbf{D}_{\text{t}}$ and normals $\mathbf{N}_{\text{t}}$:
\begin{equation}
\mathcal{L}_{\text{geo}} =  
\lambda_\text{d} \|\mathbf{D}_{\text{t}}-\mathbf{D}_\text{p}\|_1 
+ \lambda_\text{n} \|\mathbf{N}_\text{t}-\mathbf{N}_\text{p}\|_1 
+ \lambda_\text{o} \mathcal{L}_{\text{o}},
\end{equation}
where $\lambda_\text{d}$ and $\lambda_\text{n}$ are user-prescribed weights. $\mathcal{L}_{\text{o}}$ is an entropy loss on triangle opacity $\alpha$, following~\cite{guedon2024sugar}. We regularly prune triangle primitives with $\alpha < 0.5$, which removes redundant geometry and maintains a compact representation. The appearance loss supervises neural Gaussians via:
\begin{equation}
\mathcal{L}_{\text{rgb}} = (1-\lambda_{\text{c}})  \|\mathbf{C}_{\text{gt}}-\mathbf{C}_{\text{gs}}\|_{1}
+ \lambda_{\text{c}} \text{SSIM}\bigl(\mathbf{C}_{\text{gt}},\mathbf{C}_{\text{gs}}\bigr)
+ \lambda_{\text{s}} \mathcal{L}_{\text{s}},    
\end{equation}
where $\mathcal{L}_{\text{s}}$ is a volume regularization term adopted from Scaffold-GS~\cite{lu2024scaffold}. Notably, appearance gradients from Gaussians are back-propagated to the triangles, enabling implicit refinement of the underlying geometry. More details are provided in Appendix~\ref{sec:appendix_training_strategy}.

\subsubsection{Global Map Update}
In our streaming framework, camera poses are continuously refined within the backend, while primitives in the mapper are initialized and optimized using the poses available at that timestamp. This asynchronous update can lead to pose–model misalignment. To maintain consistency between the refined poses and the 3D model, we explicitly transform the primitives after the pose optimization. Specifically, we record the source keyframe for each primitive and apply a relative transformation 
$\Delta\mathbf{T} = \mathbf{T}_{\text{n}}\mathbf{T}_{\text{o}}^{-1}$ to its attributes when the corresponding keyframe pose changes from $\mathbf{T}_{\text{o}}$ to $\mathbf{T}_{\text{n}}$:
\begin{equation}
\label{Eq:global_map_adj}
\mathbf{p}'_{\text{t}} = \Delta\mathbf{T} \mathbf{p}_{\text{t}}, \quad \mathbf{o}'_{\text{g}} = \Delta\mathbf{T}(\mathbf{o}_{\text{g}} + \bm{\mu}_{\text{t}}) - \bm{\mu}_{\text{t}}', \quad
\mathbf{q}'_{\text{g}} = \mathcal{R}^{-1}(\Delta\mathbf{R} \mathcal{R}(\mathbf{q}_{\text{g}})),
\end{equation}
where $\Delta\mathbf{R}$ is the rotation component of $\Delta\mathbf{T}$, and $\mathcal{R}(\cdot)$ maps quaternions to rotation matrices. Here, $\{\mathbf{p}'_{\text{t}}, \bm{\mu}'_{\text{t}}, \mathbf{o}'_{\text{g}}, \mathbf{q}'_{\text{g}}\}$ denote the updated triangle and Gaussian parameters.
\begin{figure*}[t!]
\centering
\includegraphics[width=0.95\linewidth]{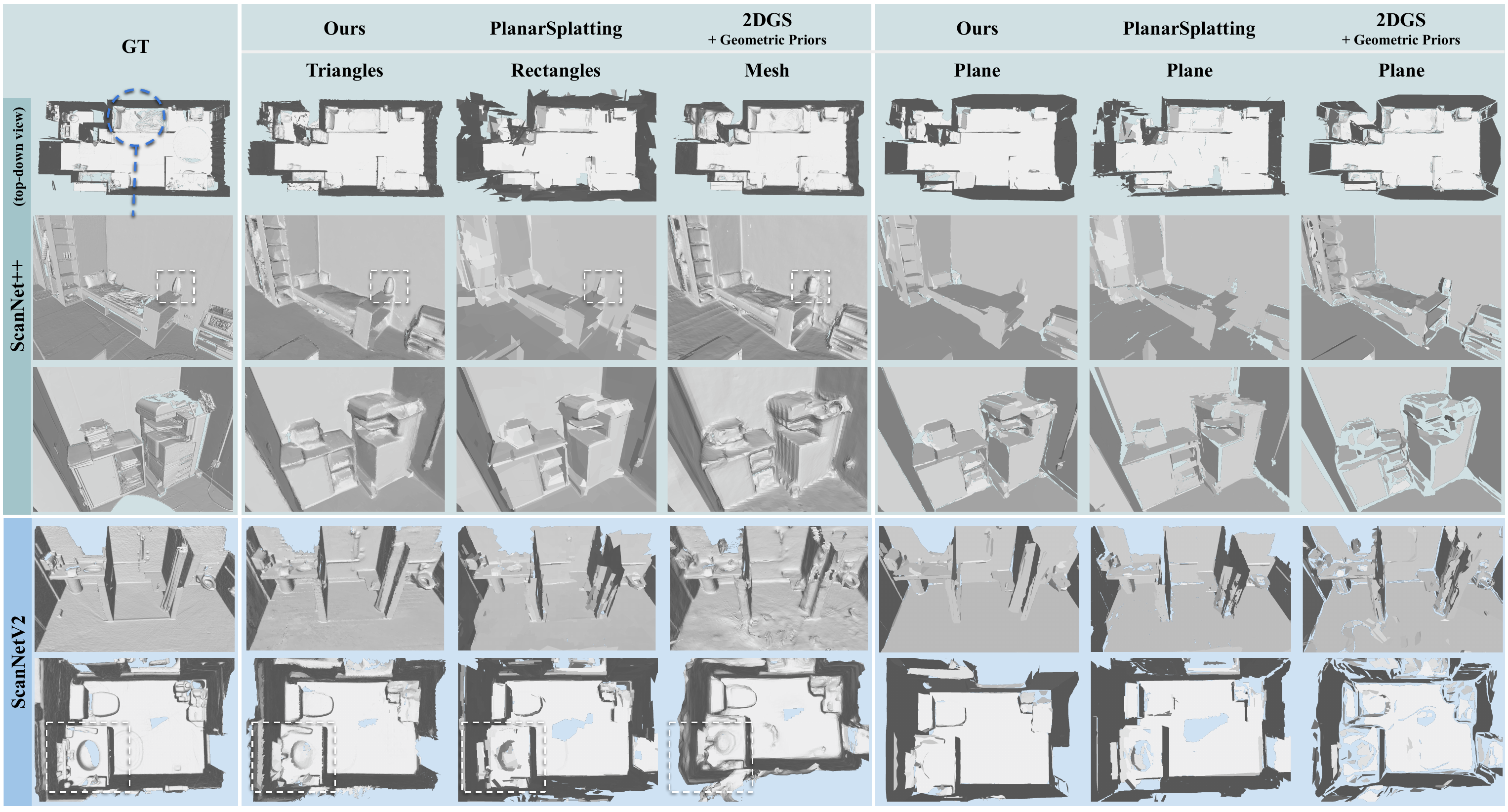}
\caption{\textbf{Qualitative comparison of geometric reconstruction.} We visualize planar reconstruction and geometric modeling across different primitives, with 2DGS shown as dense mesh for comparison. Overall, our method preserves planar structures while capturing fine geometric details.}
\label{fig:geo_compare_main}
\end{figure*}

\section{Experiments}
\label{sec:exp}

\subsection{Experimental Setup}
\textbf{Datasets.} 
We evaluate \modelname~on 56 real-world scenes from diverse benchmarks: 20 from ScanNet++~\cite{yeshwanth2023scannet++}, 10 from ScanNetV2~\cite{dai2017scannet}, 6 from VR-NeRF~\cite{xu2023vr}, 4 from FAST-LIVO2~\cite{zheng2024fast}, 8 from KITTI~\cite{geiger2012we}, and 8 from Waymo~\cite{sun2020scalability}, covering a wide range of indoor and outdoor environments.

\textbf{Baselines.}
We compare \modelname~with state-of-the-art methods across three categories. For per-scene reconstruction, we evaluate 2DGS~\cite{huang20242d}, PGSR~\cite{chen2024pgsr}, and MeshSplatting~\cite{held2025meshsplatting}. For streaming reconstruction, we select ARTDECO~\cite{li2025artdeco}, OnTheFly-NVS~\cite{meuleman2025fly}, S3PO-GS~\cite{cheng2025outdoor}, and MonoGS~\cite{matsuki2024gaussian}. For planar reconstruction, we include PlanarSplatting~\cite{tan2025planarsplatting} and AirPlanes~\cite{watson2024airplanes}. 
To ensure fair comparison, all per-scene reconstruction baselines are augmented with the same MASt3R geometric priors used in ours. 
For a controlled comparison, 2DGS, PGSR, MeshSplatting, AirPlanes, and PlanarSplatting are provided with the same camera poses estimated by \modelname, as they do not estimate camera poses within their reconstruction pipelines. Appendix~\ref{sec:pose_accuracy} shows that this shared pose input is sufficiently accurate for the reported controlled evaluation.
In contrast, streaming baselines with native tracking are evaluated using their own estimated poses.

\textbf{Metrics.} 
We conduct a comprehensive evaluation of our framework across three tasks. For planar reconstruction, following PlanarSplatting~\cite{tan2025planarsplatting}, we evaluate plane geometry using Chamfer Distance and F-score. For datasets with ground-truth plane annotations, we further assess the top-20 largest planes using Planar Fidelity, Planar Accuracy, and Planar Chamfer metrics. For dense mesh reconstruction, we report Chamfer Distance and F-score, while for novel view synthesis (NVS), we use standard metrics including PSNR, SSIM~\cite{wang2004image}, and LPIPS~\cite{zhang2018unreasonable}. In addition, we report training time and the number of primitives to quantify computational efficiency. 

\textbf{Implementation Details.}
Following standard novel view synthesis practice, we hold out every eighth frame for evaluation. Test views are used only for pose estimation and final evaluation, but are excluded from Gaussian initialization and training. Following~\cite{li2025artdeco,meuleman2025fly}, our method, ARTDECO~\cite{li2025artdeco}, and OnTheFly-NVS~\cite{meuleman2025fly} perform a 15k-iteration final global refinement over the maps constructed during streaming, whereas per-scene baselines are trained for 30k iterations. Additional implementation details are provided in Appendices~\ref{sec:implementation_details}--\ref{sec:app_geometric_prior_integration}.

\begin{figure*}[t!]
\includegraphics[width=0.95\linewidth]{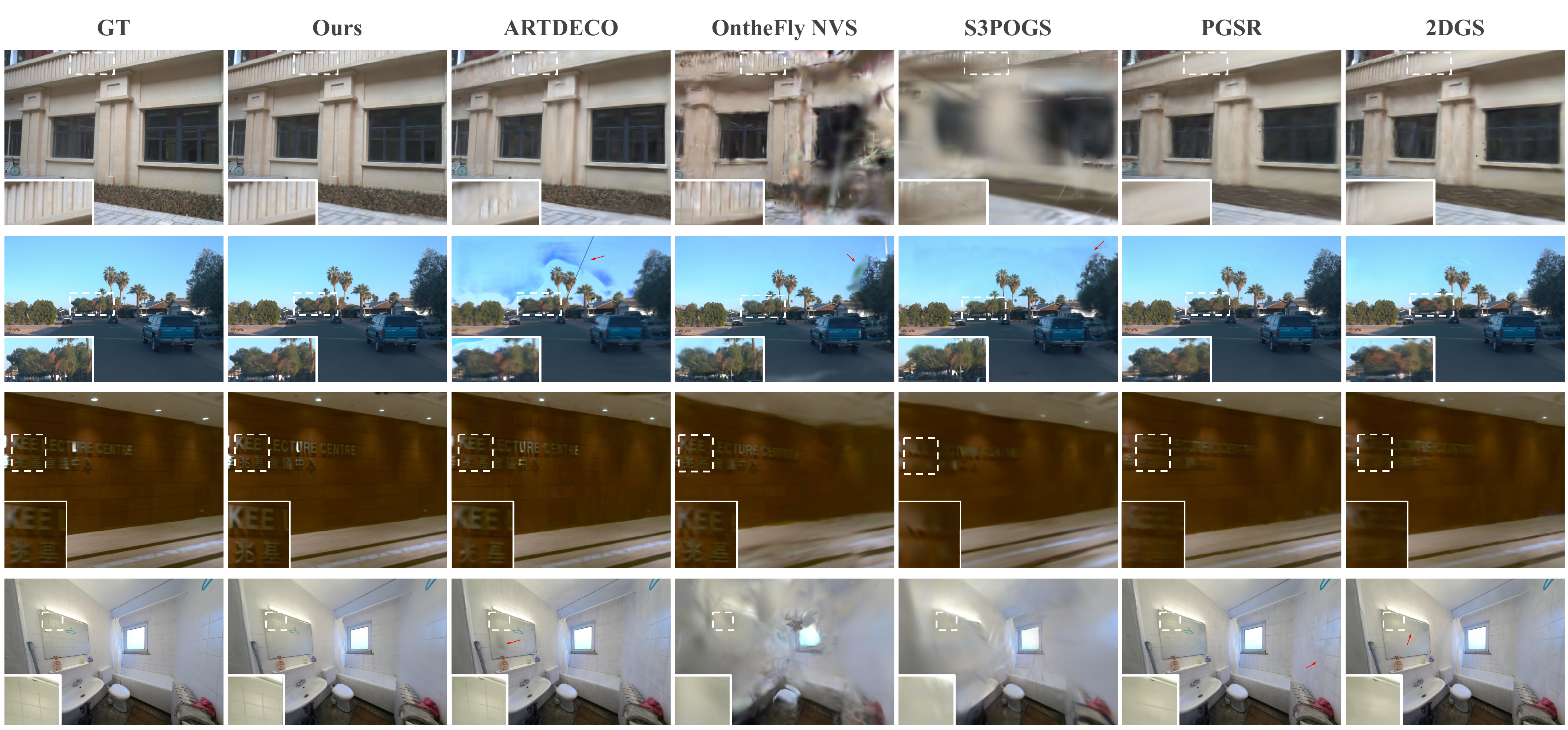}
\centering
\caption{\textbf{Qualitative comparison of appearance rendering.} We evaluate our method against state-of-the-art approaches. White wireframes highlight regions where our method excels, faithfully reconstructing fine structures and more complete surfaces.}
\label{fig:rendering_compare_main}
\end{figure*}

\subsection{Results Analysis}
\textbf{Geometry Results.}
We first evaluate planar reconstruction against six baselines covering diverse scene representations, including triangles, 3D Gaussians, surfels, rectangles, and implicit planar embeddings. As shown in Tab.~\ref{tab:plane_recon_comp}, our method consistently achieves the best accuracy, with the lowest Chamfer Distance and highest F-score, while maintaining a compact primitive count and fast training speed.
As shown in Fig.~\ref{fig:geo_compare_main}, our hybrid representation preserves planar regularity and sharp geometric features by explicitly modeling planar structures with triangles. In contrast, rectangle-based representations, despite their compactness, lack the flexibility to capture fine-grained geometry, limiting their ability to model complex scene structures. Surfel-based methods, which tightly couple geometry and appearance, often suffer from appearance-induced distortions, resulting in uneven or erroneous surfaces even when geometric priors are applied. 
Detailed dense mesh reconstruction evaluations are provided in Appendix~\ref{sec:appendix_dense_mesh}, where our method achieves consistently improved accuracy.

\textbf{Rendering Results.}
As shown in Tab.~\ref{tab:rendering_comp}, our method achieves state-of-the-art rendering performance against both per-scene optimization and streaming reconstruction baselines. It shows clear advantages in texture-less and low-light regions (Fig.~\ref{fig:rendering_compare_main}), where per-scene methods often suffer from overfitting or poor Gaussian initialization, and streaming methods are affected by pose drift. In contrast, our method leverages accurate geometry and robust feed-forward pose estimation, leading to more stable and consistent rendering quality.

\begin{figure*}[t!]
\centering

\begin{minipage}[t]{0.45\linewidth}
\centering
\vspace{0pt}

\captionof{table}{\textbf{Ablation studies on the ScanNetV2 dataset.} We conduct ablation studies on the hybrid representation and framework design, evaluating performance across both geometric and appearance metrics.}
\label{tab:ablation_comp_scv2}

\vspace{2pt}

\renewcommand{\arraystretch}{1.15}
\setlength{\tabcolsep}{0.6pt}

\resizebox{\linewidth}{!}{
\begin{tabular}{l|cc|ccc|c}
\toprule
\multirow{2}{*}{Setting} & \multicolumn{2}{c|}{Geometry} & \multicolumn{3}{c|}{Rendering} & {\# Primitives} \\
 & Ch-L2$\downarrow$ & F-score$\uparrow$
       & PSNR$\uparrow$ & SSIM$\uparrow$ & LPIPS$\downarrow$ & (\#Geo/\#GS)\\
\midrule 
Ours & \textbf{5.68} & \textbf{62.15} & \textbf{28.83} & \textbf{0.882} & 0.222 & 56.1k/222.2k\\
\midrule
w/o triangles & 5.90 & 59.85 & 28.44 & 0.876 & 0.232 & 52.8k/157.3k\\
w/o hybrid & 6.06 & 57.54 & 28.48 & 0.877 & 0.231 & -/621.5k\\
\midrule
w/o spatial filtering & 6.01 & 58.86 & 28.66 & 0.880 & \textbf{0.213} & 211.5k/625.7k\\
w/o global map update & 6.20 & 56.00 & 28.33 & 0.877 & 0.229 & 55.3k/166.6k\\
\bottomrule
\end{tabular}
}

\footnotesize{$-$: w/o geometric primitives.}

\end{minipage}
\hfill
\begin{minipage}[t]{0.5\linewidth}
\centering
\vspace{0pt}

\includegraphics[width=\linewidth]{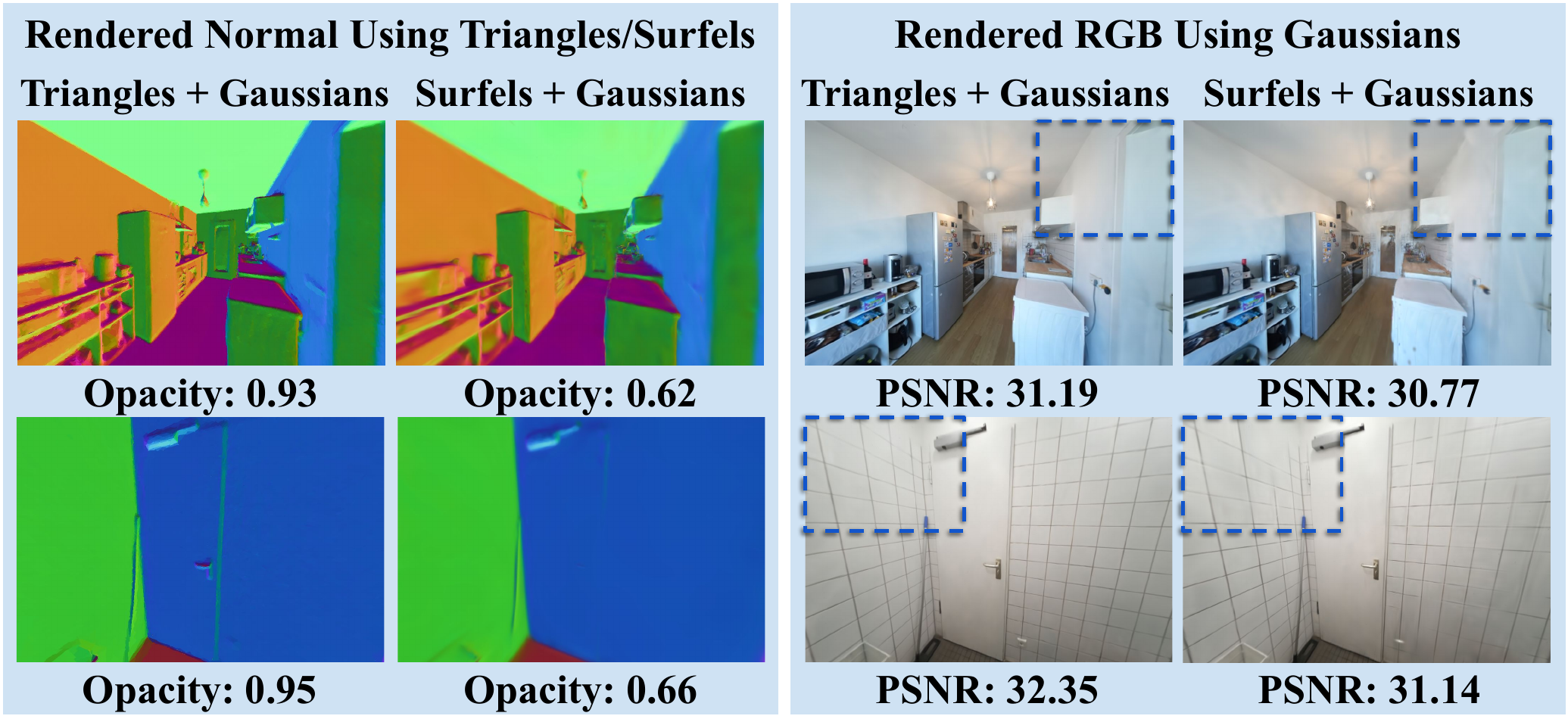}


\captionof{figure}{\textbf{Ablation on triangle representation.} 
Compared to surfels, our representation produces clearer opaque surfaces and finer rendering details.}

\label{fig:ablation_wo_triangles}

\end{minipage}
\end{figure*}

\textbf{Gain Attribution.}
Appendix~\ref{sec:app_geometric_prior_integration} (Tab.~\ref{tab:baseline_prior_effect}) isolates the effect of shared MASt3R supervision on the per-scene baselines. Although this augmentation improves the geometric accuracy of every baseline, \modelname~retains a clear advantage under the same prior protocol. \textbf{\modelname's gains therefore cannot be attributed to geometric priors alone; instead, they arise from its hybrid representation and streaming-specific map management.} The hybrid representation decouples explicit structural geometry from appearance optimization, spatial filtering controls redundant map growth, and global map updates maintain geometric consistency under pose refinement. The ablations in Tab.~\ref{tab:ablation_comp_scv2} further confirm that each component contributes to overall reconstruction performance.

\begin{figure}[!t]
\centering
\includegraphics[width=0.95\linewidth]{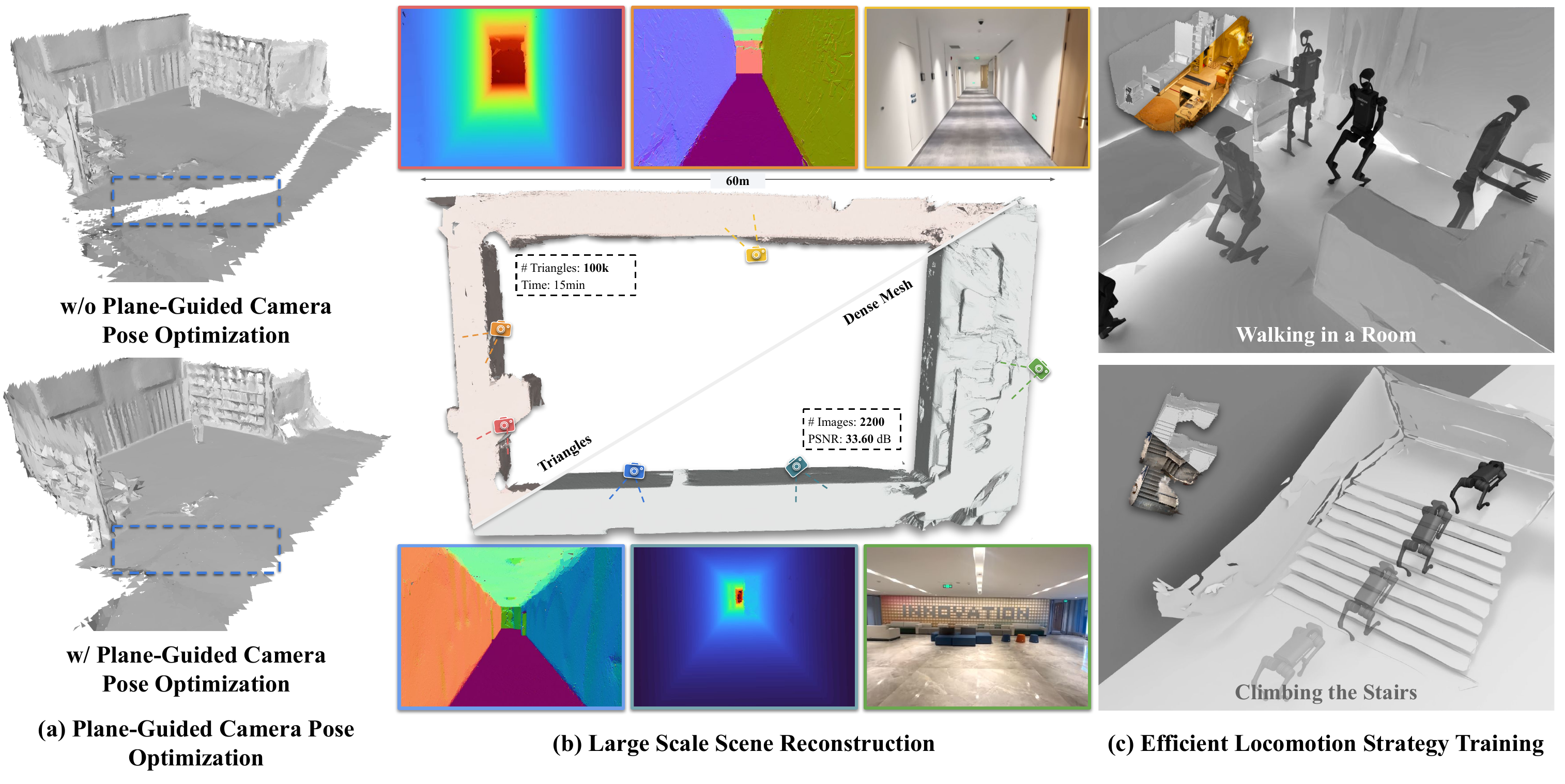 }
\caption{\textbf{Applications and system analysis.}
From left to right: 
(a) plane-guided camera pose optimization, where planar constraints effectively reduce drift; 
(b) large-scale indoor reconstruction using over 2000 monocular images; 
(c) reinforcement learning in reconstructed environments, including humanoid locomotion and quadruped stair climbing.}
\label{fig:applications_combined}
\end{figure}

\subsection{Applications}
\textbf{Plane-Guided Camera Pose Optimization.} 
Most streaming reconstruction frameworks separate pose estimation from mapping, preventing effective use of global geometry and leading to drift. We feed the resulting planar map back to the frontend and refine camera poses with a point-to-plane alignment loss, improving global geometric consistency (Fig.~\ref{fig:applications_combined}(a)). Details are provided in Appendix~\ref{sec:appendix_help_tracker}.

\textbf{Large-Scale Scene Reconstruction.}
Large-scale reconstruction remains memory-intensive even with compact primitives. We adopt a dynamic loading strategy that swaps primitive parameters between GPU and CPU, allowing the framework to scale to large environments under limited GPU memory (Fig.~\ref{fig:applications_combined}(b)). Details are provided in Appendix~\ref{sec:appendix_large_scale_scene}.

\textbf{Downstream Locomotion Demonstration.}
To assess their downstream utility, we convert the compact planar assets reconstructed by our method into physics-based simulation environments for locomotion. In the evaluated Isaac Lab tasks, their structurally consistent contact geometry supports trained policies that achieve humanoid walking and quadruped stair climbing (Fig.~\ref{fig:applications_combined}(c)). These results demonstrate the utility of the reconstructed assets for physics-based locomotion simulation. Details are provided in Appendix~\ref{sec:appendix_locomotion}.

\subsection{Ablation Studies}
We conduct ablation studies to evaluate the contributions of our representation and framework design. 

\begin{figure}[!t]
\centering
\includegraphics[width=\linewidth]{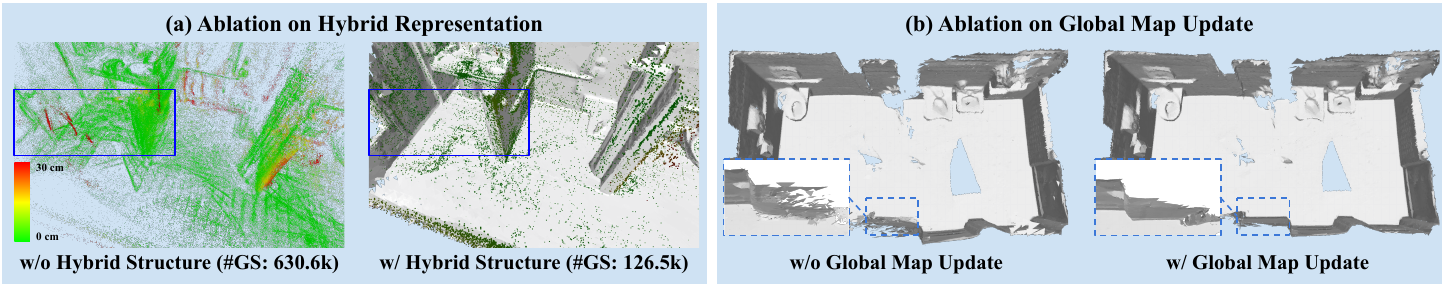 }
    \caption{\textbf{Ablation on hybrid representation and global map update.} Our hybrid design reduces representation redundancy and mitigates geometric inconsistencies in feed-forward depth, while global map updates improve overall consistency. Point clouds visualize Gaussian centers.}
   \label{fig:ablation_wo_loop_and_decoupled}
\end{figure}

\textbf{Representation Design.}
We replace triangles with surfels to ablate our triangle representation. 
Unlike surfels with soft Gaussian kernels, our sigmoid-based edge-preserving triangles provide explicit boundaries that confine gradients to structurally relevant pixels. 
This stabilizes optimization, leading to more accurate geometry and higher-quality rendering (Fig.~\ref{fig:ablation_wo_triangles}). 
We further ablate the hybrid representation by replacing it with unanchored neural Gaussians. As shown in Tab.~\ref{tab:ablation_comp_scv2}, the hybrid representation improves both geometric accuracy and rendering quality. Moreover, the proposed representation reduces redundancy and encourages Gaussians to concentrate around the underlying surface, as shown in Fig.~\ref{fig:ablation_wo_loop_and_decoupled}(a).

\textbf{Framework Design.}
We conduct ablation studies on the mapping module of our on-the-fly reconstruction framework. Disabling \textbf{spatial filtering} substantially increases the number of primitives (+200\% on ScanNetV2 and +245\% on ScanNet++), confirming its effectiveness in reducing redundancy. As shown in Fig.~\ref{fig:ablation_wo_loop_and_decoupled}(b), enabling the \textbf{global map update} improves geometric consistency and, consequently, rendering quality.
More ablation results are provided in Appendix~\ref{sec:more_ablation}.

\section{Conclusion}
\label{sec:conclusion}
\modelname~addresses a key limitation of existing streaming Gaussian-based reconstruction methods: the lack of explicit, compact, and stable geometry for anchoring appearance optimization. By introducing a loosely coupled triangle-Gaussian representation together with a streaming reconstruction framework, \modelname~decouples geometry from appearance while preserving high-fidelity rendering. This design mitigates geometric drift, redundancy, and optimization instability that arise when geometry and appearance are modeled by a single representation. Extensive results show that \modelname~enables efficient and structurally consistent streaming reconstruction, while producing compact planar assets that support downstream applications such as pose refinement, large-scale reconstruction, and simulation-ready scene modeling.


\newpage
\bibliographystyle{plain}
\bibliography{main}


\appendix 
\newpage
The appendices provide additional implementation details, application analyses, supplementary experimental results, and limitations. Specifically, Sec.~\ref{sec:appendix_technical} details the technical implementation of our method and the baseline evaluation protocol, Sec.~\ref{sec:appendix_application} describes application settings, Sec.~\ref{sec:more_results} presents supplementary experimental results, and Sec.~\ref{sec:limitations} discusses limitations.

\section{Technical Details}
\label{sec:appendix_technical}

\subsection{Differentiable Triangle Rasterizer}
\label{sec:appendix_rasterizer}
To enable unbiased depth and normal rendering with triangle primitives, we adopt an explicit ray–primitive intersection strategy~\cite{sigg2006gpu}, following 2DGS~\cite{huang20242d}.
We define the transformation from a triangle’s local coordinate system to world space as
\begin{equation}
\mathbf{H} = \begin{bmatrix}s_{u}\mathbf{t}_u&s_{v}\mathbf{t}_v&0&\bm{\mu}\\0&0&0&1\end{bmatrix},
\label{H}
\end{equation}
where $\bm{\mu}$, $s_u$, $\mathbf{t}_u$, $s_v$, and $\mathbf{t}_v$ follow the definition of the local triangle frame in Eq.~\ref{eq:local_triangle_frame}.

When combined with the edge-preserving contribution function (Eq.~\ref{eq:contribution}), this formulation leads to two practical challenges:
(i) \textbf{inaccurate depth sorting} for large triangles whose barycenters deviate from true ray–triangle intersections; and
(ii) \textbf{incorrect visibility estimation} when triangle barycenters are occluded while portions of the triangle remain visible.

\begin{figure}[ht]
\centering
\includegraphics[width=0.7\linewidth]{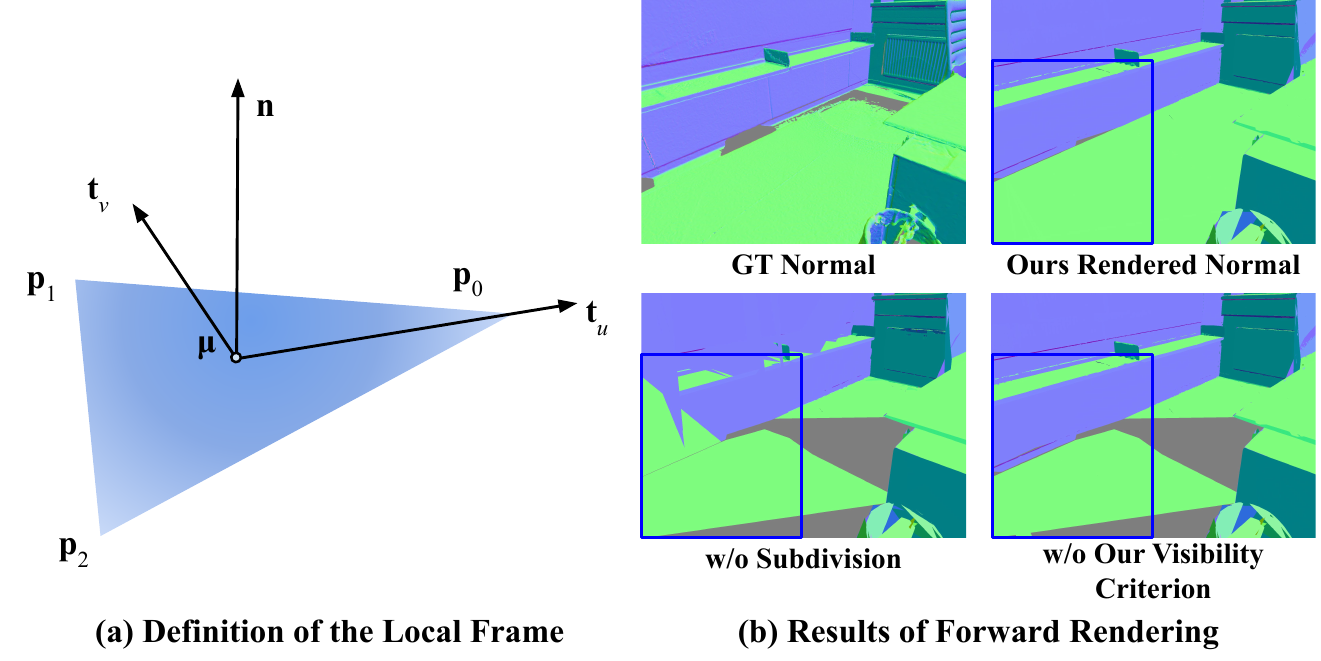}
\caption{\textbf{Definition of the local frame} and \textbf{results of forward rendering}. Our triangle rasterizer enables correct and reliable forward rendering of triangles.}
\label{fig:triangle_frame_and_rasterizer}
\end{figure}

To address these issues, we propose a subdivision-aware forward rendering pipeline that combines adaptive triangle subdivision for robust depth sorting with vertex-based visibility reasoning for accurate occlusion handling. The complete procedure is summarized in Algorithm~\ref{alg:forward_rendering}, and its effectiveness is illustrated in Fig.~\ref{fig:triangle_frame_and_rasterizer}(b).

\begin{algorithm}[!t]
\caption{Subdivision-aware Forward Rendering}
\label{alg:forward_rendering}
\KwIn{
Triangle soup $\mathcal{T}$, camera pose $\mathbf{W}$, screen resolution
}
\KwOut{
Rendered depth and normal maps
}

\textbf{Triangle Preprocessing:} \\
Initialize visible triangle set $\mathcal{T}_v \leftarrow \emptyset$ \\
\ForEach{triangle $t \in \mathcal{T}$}{
    \If{at least one vertex of $t$ is visible}{
        Construct local triangle frame and transformation $\mathbf{H}$ \;
        Subdivide $t$ recursively until all edges are shorter than threshold $\epsilon$ \;
        Assign parent triangle ID to all subdivision triangles \;
        Add subdivision triangles to $\mathcal{T}_v$ \;
    }
}

\textbf{Subdivision Processing:} \\
\ForEach{subdivision triangle $t_s \in \mathcal{T}_v$}{
    \If{at least one vertex of $t_s$ is visible}{
        Project vertices to image plane \;
        Determine overlapped tiles \;
        Compute view-space depth using barycenter of $t_s$ \;
        Generate sorting key (depth, tile ID) \;
    }
}

\textbf{Depth Sorting:} \\
Perform GPU-based radix sort on all subdivision triangles using sorting keys \;

\textbf{Rendering:} \\
\ForEach{pixel $\mathbf{x} = (x,y)^T$}{
    Define the camera ray using two orthogonal homogeneous planes\;
    Transform rays into local triangle coordinates using $(\mathbf{W}\mathbf{H})^T$ \;
    Compute ray–triangle intersection $\hat{\mathbf{x}}$ on the original triangle \;
    Evaluate rendering contribution using Eq.~\ref{eq:contribution} \;
}

Render depth and normal images following Eq.~\ref{eq:NandD} in the main text \;
\end{algorithm}

\subsection{Primitive Initialization}
\label{sec:prim_init_supp}
Upon the arrival of a keyframe from the backend, the framework determines the optimal locations for instantiating new primitives. To maintain a compact global map and mitigate structural redundancy, triangle insertion is restricted to regions exhibiting insufficient geometric coverage or high reconstruction error, guided by image-level priors. Specifically, we first apply a photometric filter, which prioritizes high-frequency regions and poorly reconstructed areas by computing an insertion probability $P_a(u,v)$ at each pixel $(u,v)$ using the Laplacian of Gaussian (LoG) operator $\Phi(\cdot)~$\cite{meuleman2025fly} to measure the discrepancy between the ground truth and rendered images as
\begin{equation}
    P_a(u,v) = \max\left( \Phi(I) - \Phi(\tilde{I}), 0 \right),
\end{equation}
where $\Phi(I) = \min(\|\nabla^{2}(G_{\sigma_g} * I)(u,v)\|, 1)$, $I$ and $\tilde{I}$ represent the ground-truth and rendered images, respectively, and $G_{\sigma_g}$ denotes a Gaussian smoothing kernel. A new geometric primitive is considered only when $P_a(u,v)$ exceeds a stochastic insertion threshold. We denote the upper bound of the sampled threshold by $\tau_a$.

\textbf{Spatial Filtering.} To suppress structural redundancy, we apply a depth-adaptive spatial filter to candidate primitives passing the photometric filter. For each candidate pixel, we compute its back-projected 3D center $\mathbf{c}_i$ and discard it if nearby geometric primitives already exist within a local vicinity, where the vicinity radius is defined in Eq.~\ref{eq:vicinity_radius}. This strategy effectively prevents excessive primitive growth in already reconstructed regions.

\textbf{Triangle Initialization.}
Once a candidate pixel $(u,v)$ is selected, a triangle is initialized.
Each triangle is parameterized by its vertices
$\{\mathbf{p}_{\text{t},k}\}_{k=0}^{2}$, opacity $\alpha_{\text{t}}$,
sharpness $\delta_{\text{t}}$, smoothness $\sigma_{\text{t}}$, and a feature
vector $\mathbf{f}_{\text{t}}$.
Following geometric scaling principles, the world-space scale is
$s_{\text{t}} = {3d_i}\big{/}{2f\sqrt{\Phi(I)}}$, where $f$ is the focal
length. The triangle orientation is determined by the normal prior at $(u,v)$.
Specifically, three unit vectors $\{\mathbf{v}_{\text{t},k}\}_{k=0}^{2}$ are
sampled on the local tangent plane at the back-projected center $\mathbf{c}_i$,
and the vertex positions are initialized as
$\mathbf{p}_{\text{t},k}=\mathbf{c}_i+s_{\text{t}}\mathbf{v}_{\text{t},k}$
for $k\in\{0,1,2\}$.
The opacity is initialized as $\alpha_{\text{t}} = 0.2\,C(u,v)$ to down-weight
low-confidence regions, where $C(u,v)$ is the backend confidence score.

\textbf{Gaussian Initialization.} Then, neural Gaussians are initialized at triangle barycenters for appearance modeling. We adaptively set the number of Gaussians per triangle to $K_{\max}$ if $\Phi(I) > 0.4$, and $K_{\min}$ otherwise. Here, the hyperparameters $K_{\max}$ and $K_{\min}$ define the bounds of the representational capacity based on scene detail. For primitive attributes, we initialize offsets $\mathbf{o}_\text{g}$, rotation $\mathbf{q}_\text{g}$, and features $\mathbf{f}_\text{g}$ to zero, while Gaussian opacity $\alpha_\text{g}$ is synchronized with $\alpha_\text{t}$. The base scale is defined as $\mathbf{s}_\text{g} = {d_i}\big{/}{2f\sqrt{\Phi(I)}} \cdot \mathbf{1}$ to align with local geometry. Crucially, the zero-order spherical harmonic coefficient $\mathbf{SH}_0$ is extracted from the pixel color at $(u, v)$, with higher coefficients zero-initialized.

\begin{table}[!t]
    \centering
    \caption{\textbf{Effect of the shared MASt3R geometric-prior integration on ScanNetV2.} The three per-scene baselines use the same \modelname-estimated poses and precomputed MASt3R priors.}
    \label{tab:baseline_prior_effect}
    \small
    \begin{tabular}{lccc}
        \toprule
        Method & MASt3R Prior & Ch-L2 $\downarrow$ & F-score $\uparrow$ \\
        \midrule
        2DGS & \ding{55} & 8.65 & 39.60 \\
              & \ding{51} & 6.48 & 53.73 \\
        \midrule
        PGSR & \ding{55} & 8.81 & 40.79 \\
             & \ding{51} & 6.59 & 54.28 \\
        \midrule
        MeshSplatting & \ding{55} & 14.63 & 23.12 \\
                       & \ding{51} & 11.15 & 30.73 \\
        \midrule
        \modelname & \ding{51} & \textbf{5.68} & \textbf{62.15} \\
        \bottomrule
    \end{tabular}
\end{table}

\subsection{Training Strategy}
\label{sec:appendix_training_strategy}
In our streaming reconstruction system, we adopt a staged training strategy to balance efficiency and reconstruction quality, following~\cite{li2025artdeco}. Specifically, when a keyframe is encountered, new primitives are initialized and the scene is optimized for $M$ iterations (set to $20$ in our implementation), while common frames are optimized for only $M/2$ iterations without adding new Gaussians. Training frames are sampled with a probability of $0.2$ from the current frame and $0.8$ from past frames to mitigate local overfitting. After processing the sequence in a streaming fashion, a global optimization is performed over all frames, prioritizing those with fewer prior updates.

\subsection{Streaming Reconstruction with Global Refinement}
\label{sec:streaming_global_refinement}
\modelname~performs incremental reconstruction as frames arrive. During this stage, the frontend estimates camera poses and selects keyframes, while the mapper continuously updates the hybrid map with newly observed scene content. The resulting map is therefore available throughout the streaming process.

After all input frames have been processed, we perform a final global refinement to improve global consistency and rendering quality. This stage refines the map constructed during the incremental reconstruction stage; it does not reconstruct the scene offline from scratch. Specifically, the refinement jointly optimizes the existing map using observations from the complete sequence, with priority given to frames that have received fewer previous updates.

Similar post-stream refinement stages are also adopted by existing online reconstruction frameworks, including OnTheFly-NVS~\cite{meuleman2025fly}, S3PO-GS~\cite{cheng2025outdoor}, ARTDECO~\cite{li2025artdeco}, SplaTAM~\cite{keetha2024splatam}, and HI-SLAM2~\cite{zhang2025hi}. The reported runtime includes this final global refinement stage.

\subsection{Planar Primitive Extraction}
\label{sec:appendeix_plane}
Planar primitives provide an efficient structural abstraction of the scene and can be directly leveraged in downstream tasks, such as robot local motion training. To extract these planes, we adopt a coarse-to-fine strategy based on GoCoPP~\cite{yu2022finding}, where the method is applied iteratively with progressively finer parameters to detect smaller planes from the residual points remaining after coarser planes are extracted.

\subsection{More Implementation Details}
\label{sec:implementation_details}
For our method, we set $K_{\min} = 4$ and $K_{\max} = 8$, with loss weights
$\lambda_\text{d} = 10.0$, 
$\lambda_\text{n} = 3.0$, 
$\lambda_\text{o} = 0.2$, 
$\lambda_\text{c} = 0.2$, and 
$\lambda_\text{s} = 0.01$.
For primitive insertion, we set the upper bound of the stochastic threshold to $\tau_a=0.5$. 
For dense mesh extraction, our method fuses triangle-rendered depth maps into meshes using TSDF, following the procedure in 2DGS~\cite{huang20242d}. 
The pose protocol and geometric-prior integration are detailed in Appendices~\ref{sec:app_pose_protocol} and~\ref{sec:app_geometric_prior_integration}, respectively.
For planar primitive extraction, since baseline methods, including 2DGS~\cite{huang20242d}, PGSR~\cite{chen2024pgsr}, MeshSplatting~\cite{held2025meshsplatting}, and ARTDECO~\cite{li2025artdeco}, typically output dense meshes, we extract multi-level planar shapes from their results using the same strategy and parameters applied to our method to ensure a fair comparison. All experiments are performed on an Intel Core i9-14900K CPU and an NVIDIA RTX 4090 GPU.

\subsection{Pose Protocol}
\label{sec:app_pose_protocol}
To ensure a controlled and fair comparison, methods that do not estimate camera poses within their reconstruction pipelines, including 2DGS~\cite{huang20242d}, PGSR~\cite{chen2024pgsr}, MeshSplatting~\cite{held2025meshsplatting}, AirPlanes~\cite{watson2024airplanes}, and PlanarSplatting~\cite{tan2025planarsplatting}, are provided with the same camera poses estimated by \modelname. This protocol controls the pose input and isolates differences in reconstruction quality. Methods with native tracking are evaluated using their own estimated poses, such that their results reflect end-to-end performance under their respective tracking and mapping pipelines.

\subsection{Geometric-Prior Integration}
\label{sec:app_geometric_prior_integration}
For the three per-scene baselines in Tab.~\ref{tab:baseline_prior_effect}, 2DGS~\cite{huang20242d}, PGSR~\cite{chen2024pgsr}, and MeshSplatting~\cite{held2025meshsplatting}, we augment each method's original objective $\mathcal{L}_{\mathrm{org}}$ with the same MASt3R~\cite{leroy2024grounding} geometric supervision:
\begin{equation}
\mathcal{L}_{\mathrm{sum}}=\mathcal{L}_{\mathrm{org}}+\mathcal{L}_{\mathrm{geo}},\quad
\mathcal{L}_{\mathrm{geo}}=
\lambda_\mathrm{d}^{\mathrm{base}}\lVert \mathbf{D}-\mathbf{D}_\mathrm{p}\rVert_1+
\lambda_\mathrm{n}^{\mathrm{base}}\lVert \mathbf{N}-\mathbf{N}_\mathrm{p}\rVert_1,
\end{equation}
where $\mathbf{D}_\mathrm{p}$ and $\mathbf{N}_\mathrm{p}$ denote the MASt3R depth and normal priors, respectively. We use $\lambda_\mathrm{d}^{\mathrm{base}}=0.2$ and $\lambda_\mathrm{n}^{\mathrm{base}}=0.1$ for these three baselines, following the geometric-prior formulation of AGS-Mesh~\cite{ren2024agsmesh}, without baseline-specific tuning. All original losses, optimization schedules, and hyperparameters are otherwise retained.

The MASt3R priors are precomputed once from the same input images and reused unchanged across these three baselines. We retain pixels with prior depth in the range $(0,100)$\,m and apply the same validity mask to both depth and normal supervision. Geometric supervision is applied throughout optimization, and all rendered depths and normals are expressed in the same camera-space convention before computing the losses.

The method-specific component is limited to obtaining the rendered depth $\mathbf{D}$ and normal $\mathbf{N}$ from each native representation. For 2DGS~\cite{huang20242d}, $\mathbf{D}$ is rendered from its 2D Gaussian surfels and $\mathbf{N}$ is computed from the rendered depth following Eq.~15 of~\cite{huang20242d} and the released implementation. For PGSR~\cite{chen2024pgsr}, $\mathbf{D}$ is its unbiased Gaussian-rendered depth and $\mathbf{N}$ is derived from depth following Eq.~5 of~\cite{chen2024pgsr}. For MeshSplatting~\cite{held2025meshsplatting}, $\mathbf{D}$ is rasterized from its triangle representation and $\mathbf{N}$ is computed from depth following its public implementation.

Tab.~\ref{tab:baseline_prior_effect} reports the effect of shared MASt3R supervision on all ScanNetV2 scenes. The same prior integration consistently improves all three baselines, reducing Chamfer-L2 by 23.8--25.2\%. These consistent gains confirm that all three baselines benefit substantially from the same augmentation, supporting the fairness of the comparison.

\section{Application Details}
\label{sec:appendix_application}

\subsection{Plane-Guided Camera Pose Optimization}
\label{sec:appendix_help_tracker}
In our streaming reconstruction system, we optionally feed back the reconstructed planar map to the frontend to refine camera poses via a point-to-plane alignment loss, improving global consistency. Specifically, in the mapper, we maintain a voxel map using a spatial hash to manage triangle primitives. During training, planar primitives are regularly extracted via region growing~\cite{hojjatoleslami1998region}. In our implementation, the voxel size is set to $3$ cm, and plane extraction is performed every $10$ frames. The extracted plane parameters and associated voxel keys are then shared with the frontend.

In the frontend, high-confidence points predicted by MASt3R~\cite{leroy2024grounding} are associated with the planar map via the voxel grid. For each point $\mathbf{p}$ and its corresponding plane, we adopt a simple yet effective point-to-plane alignment loss:
\begin{equation}
\mathcal{L}_p = \| (\mathbf{p} - \mathbf{c}) \cdot \mathbf{n} \|_1,
\end{equation}
where $\mathbf{n}$ and $\mathbf{c}$ denote the plane’s normal and center, respectively.

\begin{figure*}[t!]
\centering
\includegraphics[width=\linewidth]{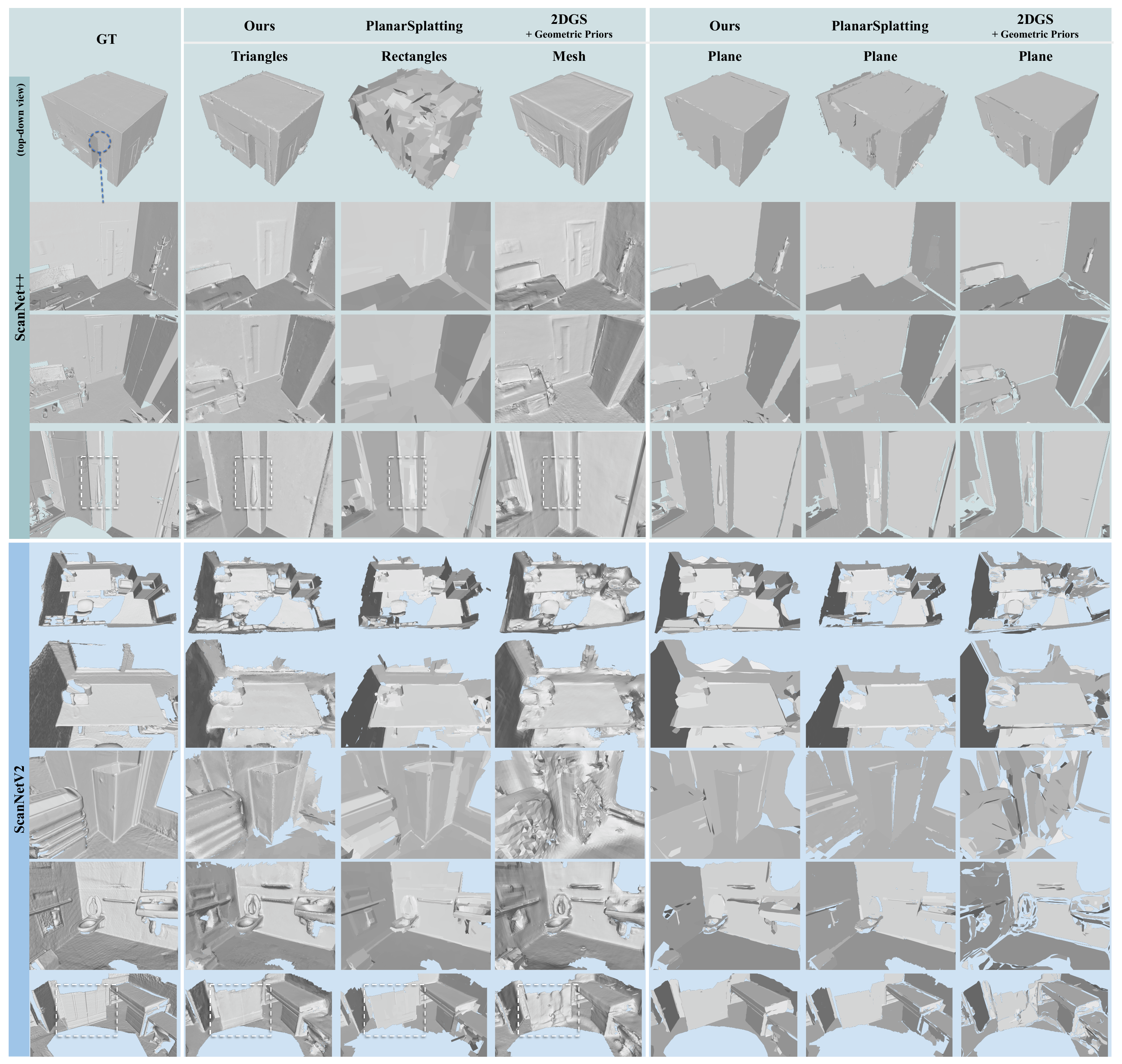}
\caption{\textbf{More geometric comparison results.} We visualize planar reconstruction and geometric modeling across different primitives, with 2DGS shown as dense mesh for comparison. Overall, our method preserves planar structures while capturing fine geometric details.}
\label{fig:geo_compare_supp}
\end{figure*}

\subsection{Large Scale Scene Reconstruction}
\label{sec:appendix_large_scale_scene}
To enable large-scale scene reconstruction and alleviate GPU memory limitations, we introduce a dynamic loading strategy that swaps primitive parameters between the GPU and CPU. Specifically, we periodically evaluate the projected scale of each neural Gaussian on the most recent image plane. A neural Gaussian is marked as \emph{invisible} if its projected scale is smaller than a pixel, and a triangle is marked as \emph{invisible} when all its associated neural Gaussians are invisible. Invisible triangles and their corresponding Gaussians are then offloaded from the GPU to the CPU. Upon detecting a loop closure in the \textit{Backend} module, we reload the primitives initialized from the associated images from the CPU back to the GPU. This dynamic loading strategy allows our framework to efficiently scale to large environments.

\begin{table}[t!]
\centering
\small
\renewcommand{\arraystretch}{1.15}
\setlength{\tabcolsep}{3pt}

\caption{\textbf{Comparison of scene import and conversion time under non-headless (GUI-based) and headless Isaac Sim pipelines.} Both settings perform the same sequence of operations, including mesh import, collision geometry construction, and USD packaging, but differ in execution mode.}
\label{tab:isaac_sim_import}

\begin{tabular}{l|c|c|c}
\toprule
Setting & Non-headless (s) & Headless (s) & \# Faces \\
\midrule
Ours (Plane) & \textbf{89.73} & \textbf{5.27} & 17k \\
\midrule
2DGS & -- & 657 & 277k \\
2DGS$^{*}$ & 120.00 & 37.21 & 17k \\
\bottomrule
\end{tabular}

\vspace{2pt}
\footnotesize{$-$: impractical runtime ($>30$ min), $^{*}$: $16\times$ mesh simplification.}

\end{table}

\subsection{Locomotion Policy Training}
\label{sec:appendix_locomotion}

Beyond visual fidelity, our hybrid representation produces compact planar geometry for downstream physics simulation. By explicitly modeling large-scale structures, such as floors and walls, with triangles, \modelname~provides structurally consistent scene assets for constructing the evaluated Isaac Lab locomotion environments.

Although 2DGS~\cite{huang20242d} is a strong reconstruction baseline, its dense meshes incur substantial preprocessing overhead in Isaac Sim. In our evaluated scene, the 2DGS mesh contains 277k faces and requires over 30 minutes for standard import and conversion. In contrast, the lower mesh complexity of \modelname~reduces processing time in both the standard and headless (\texttt{convert\_mesh}) pipelines. Quantitative comparisons are reported in Tab.~\ref{tab:isaac_sim_import}.

We evaluate the downstream utility of the reconstructed geometry through locomotion experiments in Isaac Lab using the Unitree H1 humanoid and Unitree A1 quadruped. To require policies to rely on the reconstructed geometry, we omit the height scanner from the observations. Under this setting, policies trained with aggressively simplified 2DGS meshes fail to converge because of degraded planar geometry, whereas policies trained in our reconstructed scenes achieve stable locomotion under identical observation settings. Overall, the reconstructed assets provide compact geometry for the evaluated Isaac Lab environments, demonstrating their utility for downstream locomotion simulation.

\begin{figure*}[t!]
\centering
\includegraphics[width=\linewidth]{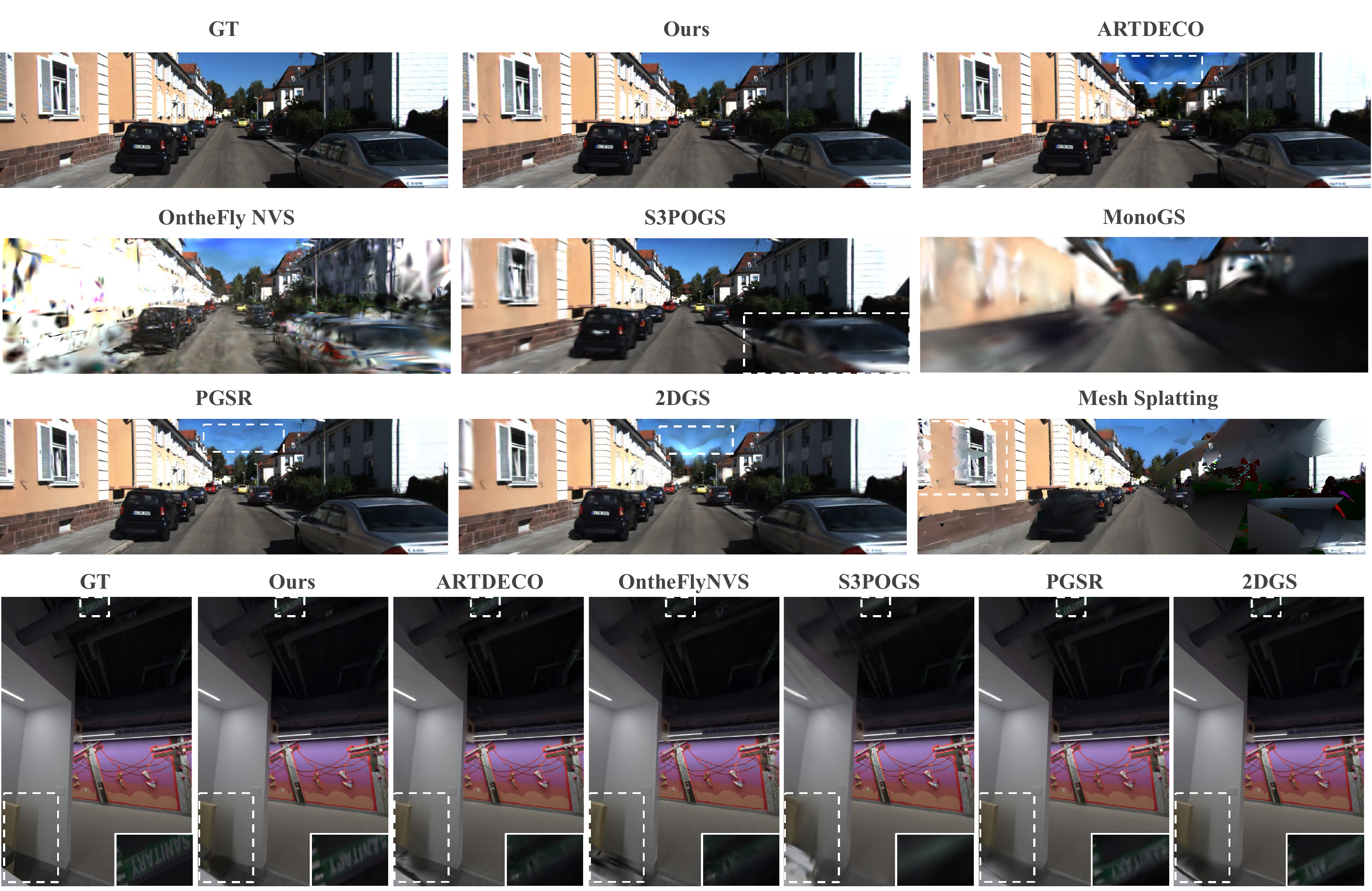}
\caption{\textbf{More rendering comparison results.} White boxes highlight artifacts and fine-grained details from baseline methods. Our approach yields significantly sharper results on intricate structures, such as text, while achieving superior overall rendering quality.}
\label{fig:render_comp_appendix}
\end{figure*}

\begin{table}[t!]
\centering
\renewcommand{\arraystretch}{1.15}
\setlength{\tabcolsep}{1pt}
\caption{\textbf{Quantitative comparison of dense mesh reconstruction.} }
\label{tab:dense_mesh_comp}
\resizebox{0.6\linewidth}{!}{
\begin{tabular}{l|cc|cc|cc}
\toprule
\multirow{2}{*}{Method} & \multicolumn{2}{c|}{ScanNet++} & \multicolumn{2}{c|}{ScanNetV2} & \multicolumn{2}{c}{FAST-LIVO2} \\
& Ch-L2$\downarrow$ & F-score$\uparrow$ & Ch-L2$\downarrow$ & F-score$\uparrow$& Ch-L2$\downarrow$ & F-score$\uparrow$ \\ 
\midrule 
2DGS$^\dagger$ & 3.95 & 80.90 & \cellcolor{yellow!40}6.45 & 53.11 & \cellcolor{yellow!40}52.83 & \cellcolor{yellow!40}61.06 \\
PGSR$^\dagger$ & \cellcolor{yellow!40}3.92 & \cellcolor{yellow!40}81.47 & 6.55 & \cellcolor{yellow!40}53.89 & 53.56 & 60.99 \\
MeshSplatting$^\dagger$ & 9.24 & 46.30 & 11.05 & 31.22 & 61.03 & 51.61 \\
\midrule
ARTDECO & \cellcolor{orange!40}3.87 & \cellcolor{orange!40}82.34 & \cellcolor{orange!40}6.00 & \cellcolor{orange!40}57.61 & \cellcolor{red!40}36.99 & \cellcolor{orange!40}61.41 \\
\midrule
Ours & \cellcolor{red!40}3.76 & \cellcolor{red!40}84.81 & \cellcolor{red!40}5.87 & \cellcolor{red!40}59.93 & \cellcolor{orange!40}38.44 & \cellcolor{red!40}64.36 \\
\bottomrule
\end{tabular}}
\begin{tablenotes}
\footnotesize
\item $^\dagger$: offline per-scene baselines are augmented with the same geometric-prior supervision as \modelname.
\end{tablenotes}
\end{table}

\section{Supplementary Experiments}
\label{sec:more_results}

\subsection{Supplementary Comparison Experiments}
\label{sec:more_comparison}
In Fig.~\ref{fig:geo_compare_supp}, we provide additional reconstruction comparisons on the ScanNet++~\cite{yeshwanth2023scannet++} and ScanNetV2~\cite{dai2017scannet} datasets, showing that our method more faithfully preserves the scene’s geometric structures. Fig.~\ref{fig:render_comp_appendix} presents rendering quality comparisons on the KITTI~\cite{geiger2012we} and VR-NeRF~\cite{xu2023vr} datasets, demonstrating our method’s robustness and applicability across diverse scenarios, including both indoor and outdoor environments.

\subsection{Dense Mesh Reconstruction}
\label{sec:appendix_dense_mesh}
We additionally evaluate the geometric quality of dense mesh reconstruction. To ensure a fair comparison, all meshes are extracted using a unified depth-fusion pipeline. Across all evaluated dense-mesh scenes, \modelname~reduces mean per-scene Chamfer-L2 by 18.52\% relative to PGSR, computed from unrounded results. As reported in Tab.~\ref{tab:dense_mesh_comp}, \modelname~achieves the lowest Chamfer-L2 on ScanNet++ and ScanNetV2, and the highest F-score across all three datasets.

\begin{table}[t!]
\centering
\renewcommand{\arraystretch}{1.15}
\setlength{\tabcolsep}{1pt}
\caption{\textbf{Ablation studies on the ScanNet++ dataset.} Our ablations are divided into two categories: representation design and framework design.} 
\label{tab:ablation_comp_scpp}
\resizebox{0.6\linewidth}{!}{
\begin{tabular}{l|cc|ccc|c}
\toprule
\multirow{2}{*}{Setting} & \multicolumn{2}{c|}{Geometry} & \multicolumn{3}{c|}{Rendering} & {\# Primitives} \\
 & Ch-L2$\downarrow$ & F-score$\uparrow$
       & PSNR$\uparrow$ & SSIM$\uparrow$ & LPIPS$\downarrow$ & (\#Geo/\#GS)\\
\midrule
Ours & \textbf{3.53} & \textbf{86.88} & 31.91 & 0.941 & 0.133 & 61.6k/291.0k\\
\midrule 
w/o triangles & 3.63 & 84.91 & 31.05 & 0.932 & 0.150 & 39.4k/201.9k\\
w/o hybrid & 3.81 & 83.01 & 31.60 & 0.940 & 0.140 & -/478.3k\\
\midrule
w/o spatial filtering & 3.71 & 84.94 & \textbf{31.95} & \textbf{0.942} & \textbf{0.126} & 233.6k/981.6k \\
w/o global map update & 3.59 & 85.07 & 31.78 & 0.941 & 0.131 & 61.5k/291.4k \\
\bottomrule
\end{tabular}}
\begin{tablenotes}
\footnotesize
\item $-$: w/o geometric primitives.
\end{tablenotes}
\end{table}

\subsection{Supplementary Ablation Studies}
\label{sec:more_ablation}
We also conduct ablation studies on ScanNet++, using the same experimental setup as in the main text. As shown in Tab.~\ref{tab:ablation_comp_scpp}, our representation design achieves the best performance in both rendering quality and geometric accuracy. Regarding system design, by enabling spatial filtering, our method can achieve higher geometric accuracy and comparable rendering quality with less than one-third of the primitives, highlighting the efficiency of our framework. Furthermore, we investigate the effect of the number of Gaussians on rendering quality. As shown in Fig.~\ref{fig:gaussian_number_ablation}, PSNR and SSIM initially improve with increasing Gaussian count and then saturate.

\begin{figure}[!t]
    \centering
    \begin{tikzpicture}

    \begin{axis}[
        width=0.53\linewidth,
        height=0.43\linewidth,
        xlabel={Number of Gaussians (k)},
        ylabel={PSNR $\uparrow$},
        xmin=100, xmax=420,
        ymin=28.5, ymax=28.85,
        xtick distance=50,
        ytick distance=0.1,
        axis y line*=left,
        axis x line*=bottom,
        tick label style={font=\small},
        label style={font=\small},
        legend style={
            font=\small,
            at={(0.97,0.03)},
            anchor=south east,
            draw=none
        },
        grid=major,
        grid style={dashed, gray!30},
    ]

    \addplot[
        color=cyan!70!blue,
        thick,
        mark=o,
        mark size=1.pt,
        mark options={fill=none},
        mark repeat=1
    ] coordinates {
        (115.178, 28.52)
        (125.3246, 28.64)
        (135.2871, 28.70)
        (166.6064, 28.83)
        (181.2875, 28.80)
        (206.7049, 28.81)
        (216.7697, 28.82)
        (226.8703, 28.79)
        (236.5143, 28.80)
        (265.1561, 28.81)
        (311.1488, 28.78)
        (357.3774, 28.81)
        (402.0284, 28.82)
    };
    \addlegendentry{PSNR}

    \addlegendimage{
        color=orange!80!yellow,
        thick,
        dashed,
        mark=o,
        mark size=1.pt,
        mark options={fill=none},
        mark repeat=1
    }
    \addlegendentry{SSIM}

    \end{axis}

    \begin{axis}[
        width=0.53\linewidth,
        height=0.43\linewidth,
        xmin=100, xmax=420,
        ymin=0.8798,
        ymax=0.8820,
        ylabel={SSIM $\uparrow$},
        axis y line*=right,
        axis x line=none,
        scaled y ticks=false,
        ytick={0.8800,0.8805,0.8810,0.8815,0.8820},
        yticklabels={0.8800,0.8805,0.8810,0.8815,0.8820},
        tick label style={font=\small},
        label style={font=\small},
    ]

    \addplot[
        color=orange!80!yellow,
        thick,
        dashed,
        mark=o,
        mark size=1.pt,
        mark options={fill=none},
        mark repeat=1
    ] coordinates {
        (115.178, 0.880175)
        (125.3246, 0.88129)
        (135.2871, 0.88114)
        (166.6064, 0.881822)
        (181.2875, 0.881291)
        (206.7049, 0.8814)
        (216.7697, 0.8814)
        (226.8703, 0.88124)
        (236.5143, 0.881119)
        (265.1561, 0.881386)
        (311.1488, 0.881607)
        (357.3774, 0.881697)
        (402.0284, 0.881578)
    };

    \end{axis}
    \end{tikzpicture}

    \caption{\textbf{Effect of the number of Gaussians on rendering quality. }PSNR and SSIM improve initially and then saturate as Gaussian count increases.}
    \label{fig:gaussian_number_ablation}
\end{figure}

\subsection{Pose Accuracy}
\label{sec:pose_accuracy}
We assess camera-pose accuracy using absolute trajectory error (ATE) RMSE, measured in meters, on ScanNetV2 and ScanNet++. Tab.~\ref{tab:pose_accuracy} compares \modelname~with representative streaming reconstruction baselines. \modelname~achieves 0.056\,m ATE on ScanNetV2 and 0.016\,m on ScanNet++. These results show that the \modelname-estimated poses are sufficiently accurate to serve as a common pose input in the controlled evaluation.

\begin{table}[!t]
    \centering
    \caption{ATE RMSE (m) on the ScanNetV2 and ScanNet++ scenes used in our reconstruction experiments. Lower is better.}
    \label{tab:pose_accuracy}
    \small
    \resizebox{0.8\linewidth}{!}{
    \begin{tabular}{lccccc}
        \toprule
        Dataset & MonoGS & S3PO-GS & OnTheFly-NVS & ARTDECO & \modelname \\
        \midrule
        ScanNetV2 & 0.246 & 0.143 & 0.289 & 0.059 & \textbf{0.056} \\
        ScanNet++ & 0.212 & 0.088 & 0.188 & \textbf{0.013} & 0.016 \\
        \bottomrule
    \end{tabular}}
\end{table}


\begin{table}[t]
    \centering
    \caption{Controlled comparison with HaloGS and GSDF on ScanNetV2 and ScanNet++. Runtime, reported in minutes, and primitive count are averaged over the two datasets.}
    \label{tab:halogs_gsdf_comparison}
    \setlength{\tabcolsep}{6pt}
    \small
    \resizebox{0.9\linewidth}{!}{
    \begin{tabular}{llccccc}
        \toprule
        Method & Dataset & Geometry Ch-L2 $\downarrow$ & Planar Ch-L2 $\downarrow$ & PSNR $\uparrow$ & Time & \# Prim.\\
        \midrule
        \multirow{2}{*}{HaloGS}
        & ScanNetV2 & 6.25 & 11.49 & 28.21 & \multirow{2}{*}{31.5} & \multirow{2}{*}{137.6k} \\
        & ScanNet++ & 3.92 & 7.76 & 31.95 & & \\
        \midrule
        \multirow{2}{*}{GSDF$^\dagger$}
        & ScanNetV2 & 7.32 & 12.39 & 28.07 & \multirow{2}{*}{85.0} & \multirow{2}{*}{369.7k} \\
        & ScanNet++ & 5.12 & 11.33 & \textbf{32.04} & & \\
        \midrule
        \multirow{2}{*}{\modelname}
        & ScanNetV2 & \textbf{5.68} & \textbf{9.07} & \textbf{28.83} & \multirow{2}{*}{\textbf{3.8}} & \multirow{2}{*}{\textbf{58.9k}} \\
        & ScanNet++ & \textbf{3.53} & \textbf{7.09} & 31.91 & & \\
        \bottomrule
    \end{tabular}}
\begin{tablenotes}
    \footnotesize
    \item $\dagger$: GSDF is augmented with the shared MASt3R depth and normal supervision described in Appendix~\ref{sec:app_geometric_prior_integration}; all other losses, training stages, and optimization settings are retained.
\end{tablenotes}
\end{table}

\subsection{Comparison with HaloGS and GSDF}
\label{sec:halogs_gsdf_comparison}

We compare \modelname~with two additional offline reconstruction methods, HaloGS~\cite{jiang2026halogs} and GSDF~\cite{yu2024gsdf}, on the ScanNetV2 and ScanNet++ datasets. The comparison evaluates geometric reconstruction, planar reconstruction, rendering quality, runtime, and representation compactness across both datasets.

For this controlled comparison, HaloGS and GSDF are provided with the same camera poses estimated by \modelname. Runtime and primitive counts are averaged over the two datasets. We retain the original training stages and full optimization budgets of both baselines. For HaloGS, which already includes a complete geometric-prior supervision scheme, we retain its original losses, weights, and optimization strategy, replacing only its prior inputs with the same MASt3R depth and normal maps used by \modelname. For GSDF, we apply the geometric-prior integration protocol described in Appendix~\ref{sec:app_geometric_prior_integration} to its geometry-producing SDF branch, using $\ell_1$ depth and normal supervision with weights $\lambda_d^{\mathrm{base}}=0.2$ and $\lambda_n^{\mathrm{base}}=0.1$, respectively. All other loss terms, training stages, and optimization settings are retained.

As shown in Tab.~\ref{tab:halogs_gsdf_comparison}, \modelname~reduces the ScanNetV2 geometric Chamfer-L2 from 6.25\,cm for HaloGS to 5.68\,cm, while requiring 58.9k rather than 137.6k primitives. It is also 8.3$\times$ faster than HaloGS and 22.4$\times$ faster than GSDF, while maintaining comparable rendering quality. These results demonstrate a favorable trade-off between geometric accuracy, representation compactness, and reconstruction efficiency.

\textbf{Compared with HaloGS, \modelname~performs 3D reconstruction in a streaming manner and maintains a persistent hybrid map as new observations arrive.} Whereas HaloGS relies on frequent triangle splitting for structural adaptation, \modelname~uses selective primitive initialization and spatial filtering to control map growth before redundant primitives accumulate, substantially improving reconstruction efficiency and representation compactness. Moreover, global map updates propagate backend pose corrections to previously constructed triangles and anchored Gaussians, maintaining map consistency throughout streaming reconstruction.

\textbf{Compared with GSDF, \modelname~uses explicit triangle primitives instead of an implicit SDF geometry branch.} This provides a direct surface representation that supports efficient incremental reconstruction while retaining high-fidelity rendering through the loosely coupled Gaussian appearance model. Together, these comparisons demonstrate the practical benefit of \modelname's streaming-oriented representation and map design.

\section{Limitations}
\label{sec:limitations}
\modelname~is a modular framework whose components can benefit from future advances in scene representation and rendering. Our current formulation inherits limitations from the chosen primitives and static-scene assumption. In particular, neural Gaussian primitives are not well suited for semi-transparent or transparent objects, where unreliable appearance gradients may adversely affect geometry optimization. Moreover, \modelname~does not explicitly model dynamic objects; inconsistent observations from moving regions may lead to erroneous primitive initialization and unstable geometry optimization. Addressing these limitations is left as future work.


\end{document}